\documentclass{article}

\usepackage[nonatbib]{neurips_2021}
\usepackage{textcomp}
\usepackage{amssymb}
\usepackage{siunitx}

\usepackage{subcaption, caption}
\DeclareCaptionStyle{ruled}{labelfont=normalfont,labelsep=colon,strut=off}

\usepackage{times}  
\usepackage{helvet}  
\usepackage{courier}  
\usepackage[hyphens]{url}  
\usepackage{graphicx} 
\usepackage{caption} 
\DeclareCaptionStyle{ruled}{labelfont=normalfont,labelsep=colon,strut=off} 
\usepackage{algorithm}
\usepackage[noend]{algpseudocode}
\usepackage{amsthm}
\usepackage{graphicx}
\usepackage{amsmath}
\theoremstyle{definition}
\newtheorem{definition}{Definition}
\newtheorem{theorem}{Theorem}
\newtheorem{lemma}{Lemma}

\DeclareMathOperator*{\argmax}{arg\,max}
\DeclareMathOperator*{\argmin}{arg\,min}
\newcommand{\tabincell}[2]{\begin{tabular}{@{}#1@{}}#2\end{tabular}}

\usepackage{newfloat}
\usepackage{multirow}
\usepackage{multicol}
\usepackage{arydshln}

\usepackage{booktabs}
\usepackage{listings}
\floatstyle{ruled}
\newfloat{listing}{tb}{lst}{}
\floatname{listing}{Listing}

\title{Federated Deep Learning with Bayesian Privacy}

\author{
Hanlin Gu $^{1,2}$
\and
Lixin Fan $^1$ \and
Bowen Li$^{4}$ \and
Yan Kang$^1$\and
Yuan Yao$^2$ \\ \And
Qiang Yang$^{1,3}$
\affiliations
$^1$AI Group, WeBank\\
$^2$Dept. of Mathematics, Hong Kong University of Science and Technology \\
$^3$Dept. of CSE, Hong Kong University of Science and Technology\\
$^4$Dept. of CSE, Shanghai Jiao Tong University
\emails
hguaf@connect.ust.hk,
\{lixinfan, brannli, yankang\}@webank.com, \\
yuany@ust.hk,
qyang@cse.ust.hk
}

\begin{document}

\maketitle

\begin{abstract}
Federated learning (FL) aims to protect data privacy by cooperatively learning a model without sharing private data among users. For Federated Learning of Deep Neural Network with billions of model parameters, existing privacy-preserving solutions are unsatisfactory. Homomorphic encryption (HE) based methods provide secure privacy protections but suffer from extremely high computational and communication overheads rendering it almost useless in practice \cite{gentry2009fully,gilad2016cryptonets,batchCryp}. Deep learning with Differential Privacy (DP) \cite{dwork2006calibrating} was implemented as a practical learning algorithm at a manageable cost in complexity\cite{abadi2016deep}.  However, DP is vulnerable to aggressive Bayesian restoration attacks as disclosed in the literature \cite{geiping2020inverting,zhao2020idlg,zhu2020deep,yin2021see} and demonstrated in experimental results of this work.

To address the aforementioned perplexity, we propose a novel Bayesian Privacy (BP) framework which enables Bayesian restoration attacks to be formulated as the probability of reconstructing private data from observed public information. 
Specifically, the proposed BP framework accurately  \textit{quantifies privacy loss by Kullback-Leibler (KL) Divergence between the prior distribution about the privacy data and the posterior distribution of restoration private data conditioning on exposed information}. To our best knowledge,  this Bayesian Privacy analysis is the first to provides theoretical justification of secure privacy-preserving capabilities against Bayesian restoration attacks. As a concrete use case, we demonstrate that a novel federated deep learning method using \textit{private passport layers} is able to simultaneously achieve \textit{high model performance, privacy-preserving capability} and \textit{low computational complexity}. Theoretical analysis is in accordance with empirical measurements of information leakage extensively experimented with a variety of DNN networks on image classification MNIST, CIFAR10, and CIFAR100 datasets. 

%

\end{abstract}

\section{Introduction}
Since the introduction of Federated learning (FL) \cite{konevcny2015federated,mcmahan2017communication}, a variety of technologies have been adopted to improve the \textit{privacy-preserving capability} and the \textit{model performance} of FL process while maintaining \textit{computational complexity} low. For instance, \textit{homomorphic encryption} (HE) \cite{gentry2009fully} provides secure protection of exchanged information and training data from being espied by semi-honest parties. However, extremely heavy computation and communication overhead incurred by HE make it unsuitable to protect {Deep Neural Network} (DNN) models which often consist of billions of model parameters. On the other hand, due to its low computational complexity, Differential Privacy (DP) \cite{dwork2014algorithmic} 
has been integrated into a practical DP-SGD algorithm to train deep neural networks with a modest privacy budget and at a manageable cost in complexity \cite{abadi2016deep}. 
Nevertheless, privacy protection provided by \textit{randomization} in DP is not as secure as that of HE.  Even with significant random noise being added, attackers may still infer training data at pixel level accuracy by launching Bayesian restoration attacks including Deep Leakage \cite{geiping2020inverting,zhao2020idlg,zhu2020deep,yin2021see,he2019model} and model inversion attacks \cite{fredrikson2015model,he2019model}.

To address aforementioned shortcomings of Homomorphic encryption and Differential Privacy, first, we propose to quantify privacy-preserving capability with a measurable Bayesian probability of reconstructing private data from observed public information. The proposed privacy loss, referred to as Bayesian Privacy (BP) loss, directly measures the removal of uncertainty about reconstructed private data, in terms of \textit{KL divergence} between \textit{the posterior belief} and the \textit{prior belief} of private data. 
To out best knowledge, BP is the first work that takes into account two adversarial processes i.e. a protection mechanism and an attack method within a unified Bayesian framework. Our experiments shown that BP losses faithfully quantify information leakage incurred by Bayesian restoration attacks. 
Second, we propose a novel Federated Deep Learning with Private Passport (FDL-PP) protection mechanism to simultaneously achieve high model performance, privacy-preserving capability and low computational complexity. Using the Bayesian Privacy framework, we also provide theoretical justification that the proposed FDL-PP method is bound to defeat various Bayesian restoration attacks. 
Extensive experiments with image classification MNIST, CIFAR10 and CIFAR100 datasets is in  accordance with theoretical analysis. 

\section{Related work} 

We review below related work from three aspects, namely, \textit{privacy definition, protection mechanism} and \textit{privacy leakage attacks}. 

\textbf{Privacy definition}: $(\epsilon, \delta)$-Differential Privacy (DP) arose as a mathematically rigorous definition of privacy for data analysis  \cite{dwork2006calibrating,dwork2014algorithmic} and since then researchers proposed a number of DP variants including \textit{moments accountants} suitable for deep learning \cite{abadi2016deep}, Gaussian Differential Privacy (GDP) with closed-form privacy bound and stronger guarantee \cite{DLGDP-Bu20} and Bayesian Differential Privacy (BDP) taking into account data distribution and achieving tighter privacy guarantees \cite{BDP-icml20}\footnote{Bayesian Differential Privacy (BDP) is different from the proposed Bayesian Privacy (BP) in that BDP as a variant of DP cannot 
effectively defense Bayesian Restoration attacks, which are formulated in the proposed Bayesian Privacy framework (see the next section)}. 


An alternative privacy definition due to \cite{eilat2020bayesian} 
proposed to measure the \textit{privacy loss} by the \textit{Kullback-Leibler (KL) Divergence} between the prior belief and the posterior belief triggered by observed messages. 
We adopt this \textit{Bayesian Privacy (BP)} analysis and generalize it as a unified framework to model both Bayesian restoration attacks and various protection mechanisms.


\textbf{Protection mechanism}: adding random noise to exchanged information i.e. \textit{randomization} is the primary protection adopted by Differential Privacy and its variants. Despite of its simplicity and popularity, this approach inevitably leads to compromised performance in terms of slow convergence, low model performance and loose privacy guarantee as documented in \cite{DLGDP-Bu20,BDP-icml20,ConvergeDP21} etc.. 
On the other hand, privacy protection provided by Homomorphic Encryption (HE) is in principle more secure than DP \cite{gentry2009fully}. However, computational and communication overhead incurred by HE are orders of magnitude higher than that of randomization approach adopted in DP. 

\textit{Split Learning} (SL) in a federated setting \cite{gupta2018distributed} proposed to separate neural network architectures into \textit{private} and \textit{public} models, and to protect private models by hiding them from other parties. This protection is effective to some extent, yet, aggressive attackers may still infer private models and data through exposed public models (as demonstrated in our experiments). 

\textit{Passport layers} was used to modulate DNN outputs for the purpose of verifying ownership and intellectual property rights (IPR) of DNN models \cite{fan2019rethinking}. We propose to integrate passport layers in private models to further enhance privacy protection capabilities. 

\textbf{Privacy leakage attacks}: it was understood that adversary of Differential Privacy seek to determine whether the observed outcome is due to one of two neighboring datasets differing by a single data point. While this \textit{hypothesis testing interpretation of Differential Privacy} was elucidated 
in \cite{StatDP-JASA10,CompDP-TIT17,InvHypoTest-PET19,HypTestDP-ICAIS20,DLGDP-Bu20}, it is limited by the neighboring datasets assumption and does not apply to aggressive gradient-based attacks recently proposed by \cite{geiping2020inverting,zhao2020idlg,zhu2020deep,yin2021see} and model inversion attacks in \cite{fredrikson2015model,he2019model}. These attacks follow the Bayesian restoration framework introduced in the present paper.

\section{Bayesian Privacy} \label{sect:BayPriv}

In this work we propose a privacy definition which is based on the notion of applying \textit{Bayesian restoration} to recover original (training) data from public information exposed during the federated learning process. Hence, the proposed framework is referred to as \textit{Bayesian Privacy} in this article. In a nutshell, \textit{Bayesian Privacy loss is quantified by Kullback-Leibler Divergence between the prior distribution about the privacy data and the posterior distribution of restoration private data conditioning on exposed information} (see Definition \ref{def:bayes privacy} for details). 

The Bayesian Privacy framework explicitly takes into account of two adversarial processes, namely \textit{protection mechanism} and  \textit{privacy leakage attacks}, each of which has their respective goals. For the sake of privacy protection, the main concern is to prevent information about original data from being disclosed i.e. to provide strong \textit{privacy guarantee}. 
The goal of \textit{privacy leakage attack}, on the other hand, is to recovery original data from exposed information e.g. the updating of deep neural network models. 

It must be noted that the modeling of these two adversarial tasks in the unified Bayesian Privacy framework is well-justified, since estimated privacy guarantee without considering leakage attacks are too loose to provide accurate accounts of information leakage. Indeed, the deficiency of Differential Privacy framework is such an example as documented in the literature \cite{BDP-icml20,DLGDP-Bu20,ConvergeDP21} etc. and demonstrated by experimental results in the present paper. 

\subsection{Bayesian Privacy in Federated learning}

In this work we consider a federated deep learning scenario, where $K$ participants collaboratively learn a multi-layered deep neural network model without exposing their private training data \cite{mcmahan2017communication, yang2019federated}. We assume certain participants or the server might be \textit{semi-honest} adversary, in the sense that they do not submit any malformed messages but may launch \textit{privacy attacks} on exchanged information from other participants.

A single client has a private data $x \in X$ and exposed public information $\mathcal{G} \in \Omega$ which is related to private data by a mapping $\mathcal{G}=g(x)$, $X \to \Omega$. 
Note that the mapping $g()$ is assumed to be \textit{known to all participants} in federated learning, while data $x$ is kept secret only to its owner. In case that $g()$ is an invertable function, it is straightforward for adversaries to infer data $ x = g^{-1}(\mathcal{G})$ from the exposed information. 

In general $g()$ is NOT invertable, 
for instance in the context of federated deep neural network learning, $g(x)$ is the exchanged neural network parameters or its gradients depending on input training data $x$.  For classification tasks, the NN model parameters $\mathcal{G}$ is sought by minimizing a loss function defined on training data $x$:
\begin{equation}
     \begin{split}\label{eq:G-loss}
        {\mathcal{G}}^{*} =  \min\limits_{\mathcal{G}}L_{CE}(\mathcal{G}, x),
        \end{split}
\end{equation}
in which $L()$ is often the \textit{cross-entropy} (CE) loss for the classification task. 

Even $g()$ in (\ref{eq:G-loss}) is NOT invertable, nevertheless, adversaries may still use $\tilde{x}$ to estimate x in a Bayesian restoration sense by minimizing $|g(\tilde{x}) - \mathcal{G}|$. Therefore, there are two  adversarial mechanisms to be considered in this framework:

\subsection{Protection Mechanism in Federated Learning}
\begin{definition} \label{def:protectM}
    A \textbf{protection mechanism} is a mapping  $\tilde{\mathcal{G}} = \mathcal{M}(\mathcal{G}), \Omega \to \Omega$ that takes as inputs exposed information $\mathcal{G}$ and outputs a modification of exposed information $\tilde{\mathcal{G}}$.
\end{definition}

    \textit{Remark}: in the literature there are various forms of protection mechanisms in federated learning 
    including, 
    \begin{itemize}
        \item \textbf{Randomization} adding a random perturbation to observed information $\tilde{\mathcal{G}} = \text{Rand}_{\epsilon}(\mathcal{G}) = \mathcal{G} +  \epsilon $, where $\epsilon $ is additive noise or perturbation. For instance, Differential Privacy \cite{geyer2017differentially} adds additive noise following Gaussian or Laplacian distribution into model or gradients; Soteria \cite{sun2021soteria}
        perturbs the output of hidden layer. 

        \item Enforcing \textbf{Sparsity} of the observed information i.e. $\tilde{\mathcal{G}} = \text{Spar}_{\{s,t\}}(\mathcal{G})$. For instance, SplitFed\cite{thapa2020SplitFed} splits a neural network model into $t$ private and $s$ public layers and only exposes partial information (public layer); Model compression \cite{zhu2020deep} prunes gradients that are below a threshold
        magnitude to leak the partial information of gradients.
        
        \item \textbf{Modulating} exposed information by adding private passport layers to DNN models, which is a novel protection mechanism parameterized by private passport $P$ and plays an indispensable role in the proposed FDL-PP method. The underlying optimization process is as follows:
        \begin{equation}
            \begin{split}
        \tilde{\mathcal{G}} = \text{Mod}_{P}(\mathcal{G}) = \argmin\limits_{\mathcal{G}} \big( L_{\text{CE}}(\mathcal{G}, x) + L_P(\mathcal{G}, P) \big),
        \end{split}
      \end{equation}
      in which $\mathcal{G}$ is model parameters in deep neural networks, $P$ is a private passport, $L_{\text{CE}}(\cdot)$ is the cross-entropy (CE) loss for classification task and $L_P(\mathcal{G}, P)$ is the regularization term to modulate $\mathcal{G}$ through the private passport $P$ (also see eq. (\ref{eq:passport-constr}) and text).
        
    \item \textbf{Homomorphic Encryption} (HE), which protects exposed information by secure encryption i.e. $\tilde{\mathcal{G}} = \text{Enc}(\mathcal{G})$ \cite{gentry2009fully}.
    \end{itemize}
    
    It must be noted that different protection mechanisms can be used in combination to provide more secure protection e.g. by adopting both \textit{splitting} and \textit{modulating} protection in the proposed FDL-PP method. Different settings of protection methods affect not only \textit{privacy} but also the \textit{model performance} of federated deep neural network models as well as the \textit{complexity} of learning process. The motivation of this research, as stated in Introduction section, is to seek appropriate protection mechanism that simultaneously achieves high privacy-preserving capability, model performance and low complexity. 
    
\subsection{Attack Mechanism with Bayesian Data Restoration in Federated Learning}
\begin{definition} \label{def:attackA}
    A \textbf{privacy leakage attack} 
    is an optimization process $\mathcal{A}$ that aims to restore the original data such that the restored data $\tilde{x}$ best fits the exposed information $\tilde{\mathcal{G}}$ i.e. 
    \begin{equation} \label{eq:leakageatt}
        \tilde{x}^* = \mathcal{A}(\tilde{\mathcal{G}}) :=  \argmin_{\tilde{x}}  {L}\large{( g(\tilde{x})},\tilde{\mathcal{G}} \large),
    \end{equation}
    in which ${L}\large( g(\tilde{x}),\tilde{\mathcal{G}} \large) $ is a loss function depending on specific attacking methods adopted by the attacker. 
\end{definition} 
The attacking in eq. (\ref{eq:leakageatt}) is derived from the Bayesian image restoration framework \cite{geman1984stochastic}, which  
dictates that inference about the true data $x$ is governed by:
\begin{equation}
   P(\tilde{x}|\tilde{\mathcal{G}}) = \frac{P(\tilde{\mathcal{G}}|\tilde{x}) P(\tilde{x})}{P(\tilde{\mathcal{G}})} \propto P(\tilde{\mathcal{G}}|\tilde{x}) P(\tilde{x}),
\end{equation}
in which $P(\tilde{x}|\tilde{\mathcal{G}}) $ is the probability of $\tilde{x}$ conditioning on the observed information $\tilde{\mathcal{G}}$, and  $P(\tilde{\mathcal{G}}|x)$ is the likelihood of observed information. 
Thus, the best restored data $\tilde{x}^*$ is sought by the optimization process:
\begin{equation} \label{eq:bayes}
\begin{split}
    \tilde{x}^* &= \mathcal{A}(\tilde{\mathcal{G}}) =\argmax\limits_{\tilde{x}}\log(P(\tilde{x}|\tilde{\mathcal{G}})) \\
    &=\argmin\limits_{\tilde{x}} [-\log(P(\tilde{\mathcal{G}}|\tilde{x}))-\log(P(\tilde{x}))] \\  &\triangleq\argmin\limits_{\tilde{x}}[{L}_1(g(\tilde{x}), \tilde{\mathcal{G}})+{L}_2(\tilde{x})]
              \end{split}
\end{equation}
in which the minimization of ${L}_2=-\log((P(\tilde{x})))$ enforces a prior of $x$, such as 
image smoothness prior by Total Variation loss $TV(\tilde{x})$ adopted in \cite{fredrikson2015model,he2019model}. 
As for the Negative Log Likelihood (NLL) loss $L_1=-\log(P(\tilde{\mathcal{G}}|\tilde{x}))$, the following forms were adopted for privacy leakage attacks:
\begin{itemize}
    \item Model inversion attack in \cite{fredrikson2015model,he2019model}
    with the loss function defined as,
    \begin{equation}
        {L}_1\large{( g(\tilde{x})},\tilde{\mathcal{G}} \large) = ||\tilde{O} - O)||^2, 
    \end{equation}
    in which $g(\tilde{x}):= \tilde{O}$ is the output of training model for the estimated $\tilde{x}$ and $\tilde{\mathcal{G}}$ is the observed output of NN parameters for the real data $x$.

    \item  Gradient-based attack method was recently studied in \cite{geiping2020inverting,zhao2020idlg,zhu2020deep,yin2021see} 
    with the loss function defined as
    \begin{equation}
        {L}_1\large{( g(\tilde{x})},\tilde{\mathcal{G}} \large) = \| \nabla W(\tilde{x}) - \big(\nabla W({x})) \|^2, 
    \end{equation}
    in which $g(\tilde{x}):=\nabla W(\tilde{x})$ is the gradients of training loss w.r.t. NN model parameters for the estimated data $\tilde{x}$, and $\tilde{\mathcal{G}}:=\nabla W({x})$ is the observed gradients of NN parameters for the real data $x$.
\end{itemize}

From attacker's point of view, the restoration of private data lead to the improvement of knowledge (or inversely the removal of uncertainty) about the private data i.e. the \textit{privacy leakage}. Suppose attackers' prior knowledge about ${\theta}$ is a distribution $F_0$, and by launching the privacy attack in eq. (\ref{eq:leakageatt}), the attacker updates the knowledge to a posterior distribution $F_{ \mathcal{A}}\big(\tilde{\theta} |\tilde{\mathcal{G}}\big)$.

\subsection{Bayesian Privacy with Adversarial Process}
\begin{definition} \label{def:bayes privacy}
Given a protection mechanism $\mathcal{M}$, a privacy leakage attack $\mathcal{A}$ and a prior knowledge distribution $F_0$ about private data $x$, the \textit{Bayesian Privacy} (BP) is defined by:
\begin{equation}
        \mathbf{BP}\big(\mathcal{M},\mathcal{A}, g, F_0 \big) = \textbf{KL}\Big(F_0 \| F_{ \mathcal{A}}\big( \tilde{x} |\mathcal{M}( g(x) ) \big) \Big),
\end{equation}
in which \textbf{KL} denotes the Kullback-Leibler Divergence between the prior knowledge distribution $F_0$ and the posterior distribution $ F_{ \mathcal{A}}$ conditioning on exposed information.
Depending on different prior knowledge $F_0$, there are two cases following the definition:
\begin{itemize}
\item \textbf{Case 1}: if attackers have zero knowledge about $x$ i.e. $F_0=\bar{F}$ is a uniform distribution over all feasible data values, \textbf{KL} divergence measures the improved knowledge about $x$ i.e. the privacy leakage loss. In this case, attacks aim to get a high \textbf{KL} values with the large knowledge increment while clients want to derive a low KL values to protect data privacy as follows,

\begin{equation}\label{eq:bayes-leak}
      \max\limits_{\mathcal{A}}\min\limits_{\mathcal{M}} \mathbf{BP}(\mathcal{M}, \mathcal{A}, g, \bar{F_0}).
\end{equation}
Consequently, we define the $\mathbf{BP_{KI}} =\mathbf{BP}(\mathcal{M}, \mathcal{A}, g, \bar{F_0}) $ as Bayesian Privacy Knowledge Increment given the prior knowledge. For any protection and attack method, we adopt $\mathbf{BP_{KI}}$ measure to quantify \textit{privacy leakage} loss of the protection mechanism $\mathcal{M}$ against the attacking method $\mathcal{A}$ with experimental results presented in the following section. \\ \textit{Remark:} If the attackers have no knowledge at first, they only randomly guess the original input, which cause the large uncertainty. They want to be sure that the restored data based on observed information is accurate instead of guessing. Specifically, adversaries want to restore the same result with multiple attacks. Some attacks in federated learning \cite{yin2021see} borrows similar idea to increase the successful rate of attack by minimizing the uncertainty of restored data with multiple attack.

\item \textbf{Case 2}: on the other hand, if the \textit{true distribution} of $x$
is known as prior to respective clients $F_0= F^*$, clients attempt to select appropriate protection mechanism such that attackers' posterior distribution is far from the true prior, and attackers aim to minimize the KL divergence,  

\begin{equation} \label{eq:BP-case2}
 \min\limits_{\mathcal{A}}\max\limits_{\mathcal{M}} \mathbf{BP}\big(\mathcal{M}, \mathcal{A}, g, F^* \big). 
\end{equation}
Note that the optimization in eq. (\ref{eq:BP-case2}) is equivalent to the minimization of the Mean Square Error (MSE) 
between the true data and restored data. This is because optimizing MSE is equivalent with optimizing KL divergence $\mathbf{KL}(F_{x^*}||F_{\tilde{x}})$, where $\tilde{x} \sim \mathcal{N}(x^*,\sigma^2 I_D)$, $x^*$ follows the true distribution.
\begin{equation}
    \begin{split}
    \mathbf{KL}(F_{x^*}||F_{\tilde{x}}) = &\mathbb{E}_{x \in P_{x^*}} \log \frac{ P_{x^*}}{P_{\tilde{x}}} =  \mathbb{E}_{x \in P_{x^*}} \log P_{(x^*|\tilde{x})}\\
    &= -\log \frac{1}{\sqrt{2\pi \sigma^2}} + \frac{1}{2\sigma^2}\mathbb{E}_{x \in P_{x^*}} (\tilde{x} - x^*)^2 
    \end{split}
\end{equation}
where $\tilde{x}=\mathcal{A}(\mathcal{M}(g(x^*)))$. \\Thus, we define $\mathbf{BP_{MSE}}(x^*, \tilde{x}) 
     =\mathbb{E}||x^*- \tilde{x}||_2^2$ as Bayesian Privacy MSE.
For any protection and attack method, we adopt the $\mathbf{BP_{MSE}}$ as Bayesian Privacy definition in this case to calculates the \textit{privacy guarantee} against possible attacks as in experimental results and Theorem presented in the following section.
\end{itemize}

\end{definition}

It must be noted that the two adversary processes \textit{protection} and \textit{attacking} in eq. (\ref{eq:bayes-leak},\ref{eq:BP-case2}) are taken separately by \textit{clients} and \textit{attackers} of federated learning respectively. The proposed Bayesian Privacy aims to provides accurate estimates of both processes within a unified framework. This theoretical analysis is in accordance with the empirical study as illustrated in the main paper and Appendix E.


\section{Federated Deep Learning with Private Passport}
This section illustrates the proposed privacy-preserving Federated Deep Learning with Private Passport (FDL-PP) and provides theoretical proof that private passport is robust against reconstruction attacks. 

\begin{figure*}[htbp]

\centering
\includegraphics[width=0.9\textwidth]{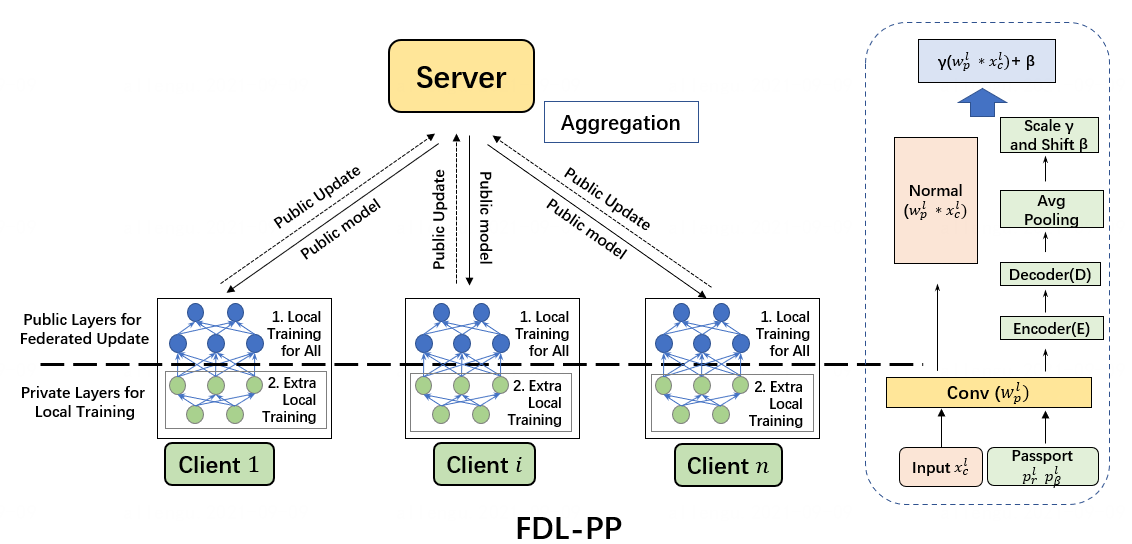}

\caption{The structure of FDL-PP. The local model is split into private layers and public layers. Only public layers are aggregated by $\texttt{FederatedAvg}$ algorithm, private layers are kept private in local(the left panel). In the private layers, the client embeds private passport into the parameters of each local model to protect the privacy of local data(the right panel).
}
\label{fig:FDL-PP}
\end{figure*}


\subsection{FDL-PP Algorithm}
Consider a neural network $\Phi(x;w,b):X \to R$, where $x\in X$, $w$ and $b$ are the weights and biases of neural networks, and $C$ is the output dimension. In the machine learning task, we aim to optimize $w$ and $b$ of network according to loss $L(\Phi(x;w,b))$,  where $x$ is the input
data and $y$ is the ground truth label. 

Inspired by \cite{thapa2020SplitFed}, we separate a deep neural network into several public ($\phi$) and private layers ($\psi$) as the 
shown in Fig. \ref{fig:FDL-PP} and the following equation:
\begin{equation} \label{eq:public_private}
    \Phi(x;w,b) = \phi^{[1]} \circ \phi^{[2]} \circ \cdot \circ \phi^{[s]} \circ \psi^{[1]} \circ \psi^{[2]}  \cdots  \psi^{[t]}(x),
\end{equation} where $s$, $t$ are, respectively, the numbers of public and private layers.
Noticeably, parameters in private layers are \textit{kept secret to preserve data privacy} and parameters in public layers are \textit{shared without encryption}.

Splitting models allows private model parameters to be kept secret, yet, training data can still be inferred from the public parameters as proved by Theorem \ref{thm:pst}.
We therefore further enhance the privacy preserving capability by adopting a private passport layer approach proposed by \cite{fan2019rethinking, zhang2020passport} to protect DNN models. 
For each private convolution layer, a private passport layer consisting of a fully connected autoencoder (Encoder $E$ and Decoder $D$) and average pooling layer is used to derive the crucial normalization parameters $\gamma, \beta$. 
Note that private passports are used to compute these parameters as follows:


\begin{equation}\label{eq:pst1}
\begin{split}
        \mathbf{O}^l(\mathbf{X}_p) = &
        \gamma^l(\mathbf{W}_p^l* \mathbf{X}_c^l) + \beta^l, \\
         \gamma^l=&\text{Avg}\Big( \mathbf{D}\big(\mathbf{E}(\mathbf{W}_p^l* \mathbf{P}_\gamma^l)\big)\Big)\\
        \beta^l = &\text{Avg}\Big( \mathbf{D}\big(\mathbf{E}(\mathbf{W}_p^l* \mathbf{P}_\beta^l)\big)\Big)
\end{split}
\end{equation}
where * denotes the convolution operations, $l$ is the layer number, 
$\mathbf{X}_c$ is the private data fed to the convolution layer. 
$\mathbf{O}()$ is the corresponding linear transformation of
outputs, while $\mathbf{P}^l_\gamma$ and $\mathbf{P}^l_\beta$
are \textit{passports private to each client} used to derive scale factor $\gamma$ and bias term $\beta$ respectively. 

The right part of Fig. \ref{fig:FDL-PP} illustrates the procedure of passport embedding. Since both the private layer parameters and passport are kept secret, it is impossible for adversary to infer training data from the private layers' outputs. In addition, we observe the sparsity of weights after embedding passport, which prevents attackers to restore the training data (see proof in Theorem \ref{thm:pst})

The training procedure of FDL-PP is illustrated in following three steps (shown in Algorithm \ref{alg:fdl-pp}):
\begin{itemize}
    \item Firstly, all clients carry out the forward and backward propagation on a global model in parallel and upload their 
    public layers of the model to the server;

    \item Secondly, the central server process the FederatedAverage \cite{mcmahan2017communication} of the public layers and update;
    
    \item Finally, clients download the updated public model, followed by back-propagation and updating to their  private layers with embedding the passport.
\end{itemize}


\begin{algorithm}[t]
	\caption{FDL-PP Framework}
	\begin{algorithmic}[1]
	   \Statex \textbf{Input:} communication rounds $T$, client number $N$, number of local epochs $E$, each client $i$ with its own passport tuple $(P_\gamma, P_\beta)$ 

        \Statex \textbf{Output:} the final model $\Phi$ \vspace{4pt}
	    
	    \State Initialize $W_0$
		\For{$t$ in communication round $T$} \vspace{2pt}
		\State Send global model parameters $W_t$ to each clients
		\For{$i$ in clients $N$} \textbf {in parallel}
		\State $W_i^{t} \leftarrow$ \textbf{Local Training}
	    \State Send public layer update $Pub(W_i^{t})$
	    \State $Priv(W_i^{t}) \leftarrow$ \textbf{Local Private Training}
		\EndFor 
		\State \textbf{Server Update:}
        \State Aggregate $\{Pub(W_t^i)\}_{i=1}^{N}$ with $\texttt{FedAvg}$ algorithm
		\EndFor 
		
		\State \textbf{Local Training:}
        
		\State  Initialize passport keys  $(P^l_\gamma, P^l_\beta)$ for targeted passport layers $N_{pass}$

		\For{number of training iterations}
			\State Sample mini-batch $(X, Y)$
			\For{$l$ in $N_{pass}$}
				\State Cross-entropy loss $L=L(X;P^l_\gamma, P^l_\beta; Y)$
			\EndFor
		
			\State Backpropagate using $L$ and update $W_i^{t}$
		\EndFor

		\State \textbf{Local Private Training:}
        \For{number of training iterations}
            \State Sample mini-batch $(X, Y)$
            \State Compute cross-entropy loss $L=L(X;P^l_\gamma,P^l_\beta; Y)$
			\For{layer $l$ in Public layers set $L_{pub}$}
				\State Frozen model parameters in layer $l$
			\EndFor
			\State Backpropagate using $L$ and update $Priv(W_i^{t})$	
		\EndFor
	\end{algorithmic}\label{alg:fdl-pp}
\end{algorithm}

\subsection{Analysis of Privacy Preserving with Private Passport}
Consider a neural network $\Psi(x;w,b,\tilde{w}, \tilde{b}): \phi \circ \psi(x)$ including one public ($\phi$) and one private layer $\psi$ as Def. \ref{def:SplitFed}, where $x \in \mathbb{X}$, defining a mapping from $\mathbb{X}$ to $R^c$, c is output dimension. In the neural network, we optimize $w$,$b$ through a loss function $L(\Psi(x;w,b,\tilde{w}, \tilde{b}), y)$, where $x$ is input data, $y$ is ground truth labels.

\begin{definition} \label{def:SplitFed}
Define the forward function of public ($\phi$) and private layer $\psi$: 
\begin{itemize}
    \item For private layer: $O = \psi(x) = a(D_\gamma \cdot w\cdot x + b)$.
    \item For public layer: $\mu = \phi(O) = a(\hat{D_\gamma} \cdot \hat{w} \cdot O + \hat{b})$.
\end{itemize}
where $a$ is activation function; $w$, $\hat{w}$ are 2D matrix of fully-connected weights of private and public layer; $b$, $\hat{b}$ are bias vector; $D_\gamma$, $\hat{D_\gamma}$ are diagonal matrix of scaling weights of batch normalization or passport layer; $\mu$, $O$ are the outputs of public and private layer; $\cdot$ denotes matrix product.

\textit{Remark}: Any convolution layers can be converted into a fully-connected layer by simply stacking together spatially shifted convolution kernels (\cite{ma2017equivalence}). In addition, the scaling factor $\gamma$ of batch normalization or passport layer could be explained by a diagonal matrix in the forward function. Consequently, we use $a(D_\gamma \cdot w \cdot x + b)$ as forward function of private layer.
\end{definition}

In the federated learning, the models or gradients of the public layers are leaked to server and other clients while are not leaked for private layers. \textit{Semi-honest} adversary participants or the server aims to restore $x$.
Theoretical analysis of privacy-preserving capabilities concerns two aspects: (1) it is easy for attackers to restore the similar input compared to original input even if the private layers are not known in the SplitFed; (2) embedding passport into the private layers 
prevents the attackers to restore the original data. 

\begin{lemma} [\cite{fan2020rethinking}] \label{lemma:public_grad}
If Partial derivative $\bigtriangledown_{\hat{b}}L(\phi(O;\tilde{w},\tilde{b}), y)$ is non-singular, $O$ could be reconstructed by solving
\begin{equation}
    \bigtriangledown_{\hat{b}}L(\phi(O;\hat{w},\hat{b}), y) = \bigtriangledown_{\hat{w}}L(\phi(O;\hat{w},\hat{b}), y) \cdot O^T
\end{equation}
\end{lemma}
\textit{Remark}: The lemma \ref{lemma:public_grad} provides that the input of public layer or output of the private layer could be easily attacked by solving the linear equations based on the leaked public model gradients and weights. Therefore, we will mainly provide the analysis of private layer given the output of private layer (input of public layer).

\begin{definition}\label{def:attack setting}
Denote $W=D_\gamma \cdot w$ and $\tilde{W} = \tilde{D_\gamma} \cdot \tilde{w}$, and $W, \tilde{W}\in$ $C^{m\times n}$($m \geq n$), an attack estimate the original private data, $x = W^+(O)$ by $\tilde{x} = \tilde{W}^+(O)$ 
where $O\in C^{m\times 1}$ is known to all participants according to Lemma \ref{lemma:public_grad} including the attacker, $+$ denote the pseudo inverse of matrix.
\end{definition}

\begin{lemma} (Wedin's Theorem \cite{stewart1990matrix})
Let $W$, $\tilde{W}$ $\in$ $C^{m\times n}$($m \geq n$). If $rank(W) = rank(\tilde{W})$, then $||\tilde{W}^+ -W^+ || \leq C ||W^{+}||_2||\Tilde{W}^{+}||_2 ||\tilde{W}-W||_2$, where $C$ is a constant.
\label{lemma:wendi}
\end{lemma} 


\begin{lemma} \label{lemma:lemma3}
Let $x=W^{+}O$ and $\Tilde{x}=\tilde{W}^{+}\Tilde{O}$, where $W$, $\tilde{W}$ $\in$ $C^{m\times n}$($m \geq n$). If $rank(W) = rank(\tilde{W})$ then
$||\tilde{x}-x|| \leq C (||W^{+}||_2||\Tilde{W}^{+}||_2||O||_2 ||\tilde{W}-W||_2 + ||W^+||_2||O-\tilde{O}||_2$
\begin{proof}
\begin{equation}
    \begin{split}
        ||x-\tilde{x}|| &= ||W^+O-\tilde{W}^+\tilde{O}|| \\
        &  \leq || W^+ \cdot O  -\tilde{W}^+\cdot O + \tilde{W}^+\cdot O - \tilde{W}^+ \cdot \tilde{O}|| \\
        &\leq ||W^+ \cdot O  -\tilde{W}^+\cdot O|| + ||\tilde{W}^+\cdot O - \tilde{W}^+ \cdot \tilde{O}|| \\
        &\leq C( ||\tilde{W}^{+}||_2||W^{+}||_2||W-\tilde{W}||_2||O||_2 \\
        & \quad + ||\tilde{W}^+||_2||O-\tilde{O}||_2)
    \end{split}
\end{equation}
The last step is because Lemma \ref{lemma:wendi}, where $C$ is a constant.
\end{proof}
\end{lemma}

\begin{lemma} \label{lemma:lemma4}
Let $x=W^{+}O$ and $\Tilde{x}=\tilde{W}^{+}\Tilde{O}$, where $W$, $\tilde{W}$ $\in$ $C^{m\times n}$($m \geq n$). If $rank(W) > rank(\tilde{W})$, then for unitary vector $O \in R(W)$ that is orthogonal to $R(\tilde{W})$ or $O \in R(W^T)$ that is orthogonal to $R(\tilde{W^T})$, where $R()$ is the row space of the matrix, we have
\begin{equation}
    ||W^+O- \tilde{W}^+ O||_2 \geq \frac{1}{||W -\tilde{W}||_2}
\end{equation}
\begin{proof}
Since unitary vector $O \in R(W)$ that is orthogonal to $R(\tilde{W})$ or $O \in R(W^T)$ that is orthogonal to $R(\tilde{W^T})$, 
\begin{equation}
    \begin{split}
        1 &= O^T O = O^TWW^TO \\
        & = O^T(\tilde{W} + W -\tilde{W})W^TO\\
        & = O^T( W -\tilde{W})W^TO \\
        &\leq ||W -\tilde{W}||_2 ||W^TO||_2
    \end{split}
\end{equation}
Hence $||W^TO||_2 \geq \frac{1}{||W -\tilde{W}||_2}$. \\
    Note that $\tilde{W}^{+}O = 0$, thus,
    \begin{equation}
        ||(\tilde{W}^T-W)O|| = ||W^TO||_2 \geq \frac{1}{||(\tilde{W}^+-W)||_2}
    \end{equation}

\end{proof}
\end{lemma}

\begin{theorem}
 \label{thm:pst}
Suppose private layer setting as the definition \ref{def:SplitFed} and \ref{def:attack setting} illustrated, and $a$ is the activation function.
\begin{itemize}
    \item \textbf{Case 1} (No passport) If inverse function $a$ satisfies Bi-Lipschitz condition and $rank(W)=rank(\tilde{W})$, 
    then
    \begin{equation} \label{eq:SplitFed}
        0 \leq \sqrt{\mathbf{BP_{MSE}}(x, \tilde{x})} \leq C_1||\tilde{W}^{+}||_2||W^{+}||_2||W-\tilde{W}||_2
    \end{equation}

\begin{proof}
Without loss of generality, we assume $b$ and $\tilde{b}$ equal to 0 because $W\cdot x + b = [W, b] \cdot [x^T, 1]^T$, and assume $O$ is unitary vector. Set $W \cdot x = O_1$, $\tilde{W} \cdot \tilde{x} = O_2$.
 Since activation function $a$ satisfies Bi-lipschitz condition and invertible, $||\tilde{W}\cdot \tilde{x}  - W \cdot x || \leq C_1||\tilde{O}-O||_2$ and $||W\cdot x||$ is bounded by $C_1||O||_2$.\\
 According to the lemma \ref{lemma:lemma3}:
 \begin{equation}
 \begin{split}
          \sqrt{\mathbf{BP_{MSE}}(x, \tilde{x})}&=||\tilde{x} -x||_2 \\
          &= ||W^+\cdot O_1 - \tilde{W}^+ \cdot O_2 ||\\ 
          &\leq C_1||\tilde{W}^{+}||_2||W^{+}||_2||W-\tilde{W}||_2
\end{split}
 \end{equation}
 where $O_1 = O_2$ as the output of private layer is fully restored.
\end{proof}
\textit{Remark}: In particular, when applying into more private layers, we could draw the similar conclusion by induction. \\
\textit{Remark}: if no passport layers are used in the private NN model, the private model between different clients often learn similar weights. 
A semi-honest attacker may take advantage of this characteristic and replace other participants' model with his/her own model.  
Eq. \ref{eq:SplitFed} then shows that the restoration attack based on surrogate models can be very successful, since the Bayesian Privacy guarantee measured by Mean-Squared-Error (MSE) might be be down to a small value provided that differences between parameters of the private model and the attacker's surrogate model is zero, which is empirically justified by experimental results shown in next section. 
    
    \item \textbf{Case 2} (Embedding passport) If $rank(W) > rank(\tilde{W})$ caused by embedding passport and $a$ satisfies Lipschitz condition and is invertiable, for $a^{-1}(O) \in R(W)$ that is orthogonal to $R(\tilde{W})$ or $a^{-1}(O) \in R(W^T)$
    that is orthogonal to $R(\tilde{W^T})$, we have
    \begin{equation}\label{eq:pst}
    \frac{C_2}{||D_\gamma||_2+\|\tilde{D}_\gamma\|_2} \leq \mathbf{BP_{MSE}}(x, \tilde{x}) \leq +\infty
        \end{equation}
    \begin{proof}
    Since $a$ satisfies lipticiz condition and invertiable, 
    $a^{-1}(O)$ exits and bounded. With loss of generality, suppose $a^{-1}$ is a unitary vector. Because $a^{-1}(O) \in R(W)$ that is orthogonal to $R(\tilde{W})$ or $a^{-1}(O) \in R(W^T)$ that is orthogonal to $R(\tilde{W^T})$, \\

    thus, according to Lemma \ref{lemma:lemma4},
    \begin{equation}
        \begin{split}
             \mathbf{BP_{MSE}}(x, \tilde{x})&=||\tilde{x} -x|| \\
             & = ||W^+O -\tilde{W}^+O||_2 \\
             & \geq \frac{1}{||\tilde{W} -W||_2} \\
             & \geq \frac{1}{||\tilde{W}||_2 + ||W||_2} \\
             & = \frac{1}{||\tilde{D_\gamma} \cdot \tilde{w}||_2 + ||D_\gamma \cdot w||_2} \\
             & \geq \frac{1}{||\tilde{D_\gamma}||_2||\tilde{w}||_2 + ||D_\gamma||_2 ||w||_2}\\
             & \geq \frac{C_2}{||D_\gamma||_2+\|\tilde{D}_\gamma\|_2}
        \end{split}
    \end{equation}
    The last inequality is right as the convolution weights ($w$ and $\tilde{w}$) bounded by constant $\frac{1}{C_2}$
    \end{proof}
    \textit{Remark}:Leaky Relu, one of Activate function satisfy bi-lipticiz condition and invertiable.
    
    \textit{Remark}: if private passport layers are used in the NN model, due to the private passports employed by different participants, 
    scaling weights in diagonal matrix $||D_\gamma||_2+ ||\tilde{D}_\gamma||_2$ decrease. Subsequently, the lower bound in eq. (\ref{eq:pst}) become larger and provide strong Bayesian Privacy guarantee measured by Mean-Squared-Error against restoration attacks. Moreover, embedding passport cause the sparsity of scaling factor of the passport layer such that the rank of $W$ drop seriously. (All experimental verification details see in next section). 
\end{itemize}
\end{theorem}

\section{Experimental Results}

This section illustrates empirical evaluation of the proposed privacy-preserving method (FDL-PP)  against existing solutions using the proposed \textit{Bayesian Privacy} framework and experimental verification for the assumptions of Theorem \ref{thm:pst}. 

\subsection{Experiment Setup and Evaluation Metric}

We investigate image classification problems on \textit{three datasets} i.e. MNIST, CIFRA10 and CIFAR100 by utilizing three backbone networks (LeNet, AlexNet and ResNet-18). 
All experiments settings are briefly summarized as follows and elaborated in Appendix. 

\textbf{Methods}:  we compare the proposed FDL-PP method against four existing federated deep learning methods, namely, Distributed Deep Learning without privacy protection (\textbf{DL}) \cite{mcmahan2017communication}, Deep Learning with Differential Privacy (\textbf{DLDP}) \cite{abadi2016deep} applying into federated learning \cite{geyer2017differentially}, \textbf{SplitFed} \cite{gupta2018distributed,thapa2020SplitFed} and Federated Deep Learning with two Homomorphic Encryption variants \textbf{CryptoNet} \cite{gilad2016cryptonets} and \textbf{BatchCrypt} \cite{batchCryp}.  Note that there are two variants of DLDP method i.e.  DLDP-A and DLDP-B respectively corresponding to small ($\sigma=0.002$) and large ($\sigma=0.128$) Gaussian noise added. 
The proposed \textbf{FDL-PP} method is implemented by splitting three backbone networks, keeping  the first convolution and activation layers as the private layer, the rest layers as the public layer. The passport layer is appended after the private layer. See Fig. \ref{fig:FDL-PP} and Algorithm \ref{alg:fdl-pp} for method details. 

\textbf{Evaluation metrics}: we compare \textit{model performances} of different methods using classification \textit{actuaries} on testing dataset. For learning \textit{complexity} we report \textit{elapse time} in seconds for one epoch of each learning method. 

For \textit{privacy-preserving} capabilities, we adopt the proposed Bayesian Privacy ($\mathbf{BP_{KI}}$ and $\mathbf{BP_{MSE}}$) defined by Def. \ref{def:bayes privacy}. In particular, two attacking methods i.e. \textbf{Deep leakage attack:DLG}( \cite{zhu2020deep} as one of the gradient-based attack and \textbf{Model Inversion attack:MI} \cite{he2019model} were employed 10 times to attack different learning methods. Corresponding Bayesian Privacy  ($\mathbf{BP_{KI}}$ and $\mathbf{BP_{MSE}}$) were computed using the posterior distribution of restored private data.

\subsection{Comparison of Privacy Preserving Mechanisms}

\begin{figure}[h!]
  \centering
  \hspace{-9.8mm}
  \begin{subfigure}{0.2\textwidth}
  \centering
    \includegraphics[width=46pt]{{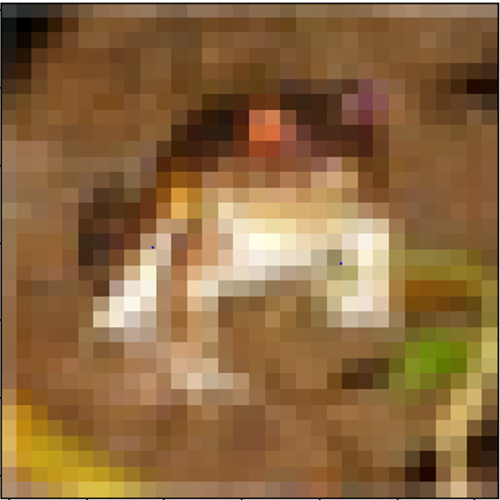}}
    \subcaption*{\textbf{$\text{BP}_\text{{MSE}}$}:0}
  \end{subfigure}
  \hspace{-10mm}
  \begin{subfigure}{0.2\textwidth}
  \centering
    \includegraphics[keepaspectratio=true, width=46pt]{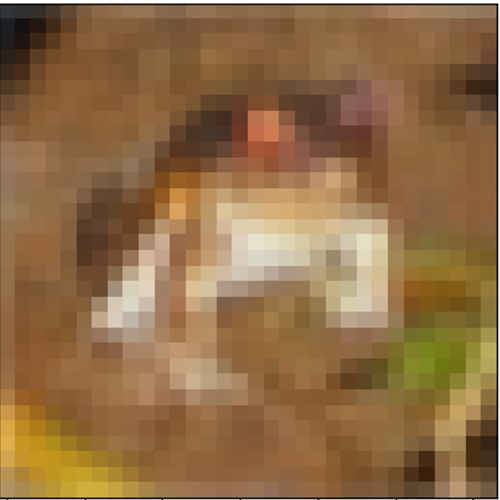}
    \subcaption*{$2.2\text{e-}8$}
  \end{subfigure}
    \hspace{-10mm}
  \begin{subfigure}{0.2\textwidth}
    \centering
    \includegraphics[keepaspectratio=true, width=46pt]{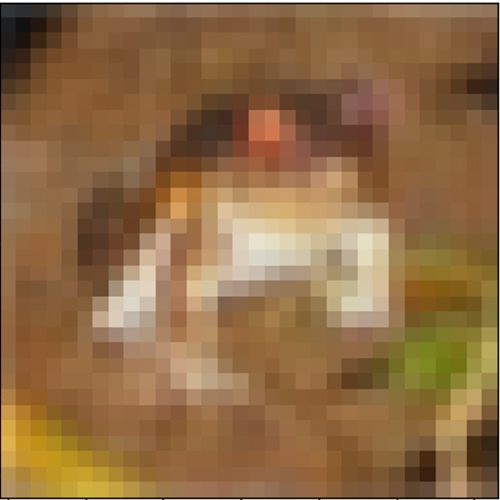}
    \subcaption*{$1.4\text{e-}6^{\Huge{*}}$}
  \end{subfigure}
    \hspace{-10mm}
  \begin{subfigure}{0.2\textwidth}
    \centering
    \includegraphics[keepaspectratio=true, width = 46pt]{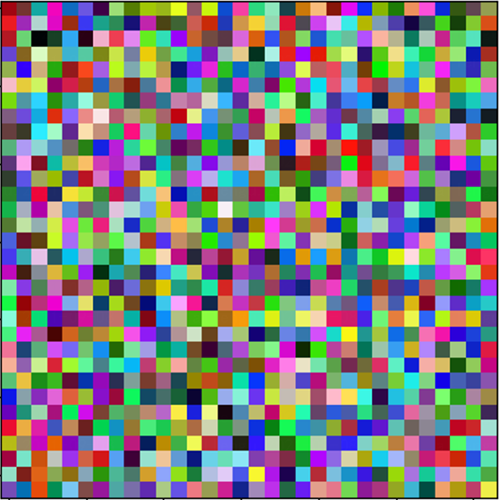}
    \subcaption*{$5.3\text{e-}2$}
  \end{subfigure}
    \hspace{-10mm}
  \begin{subfigure}{0.2\textwidth}
    \centering
    \includegraphics[keepaspectratio=true, width=46pt]{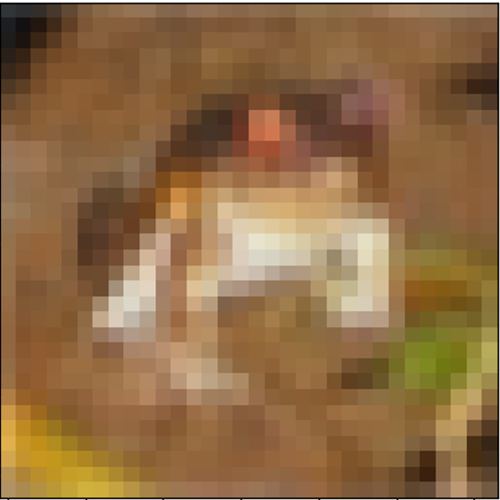}
    \subcaption*{$8.1\text{e-}5$}
  \end{subfigure}
    \hspace{-10mm}
  \begin{subfigure}{0.2\textwidth}
    \centering
    \includegraphics[keepaspectratio=true, width=46pt]{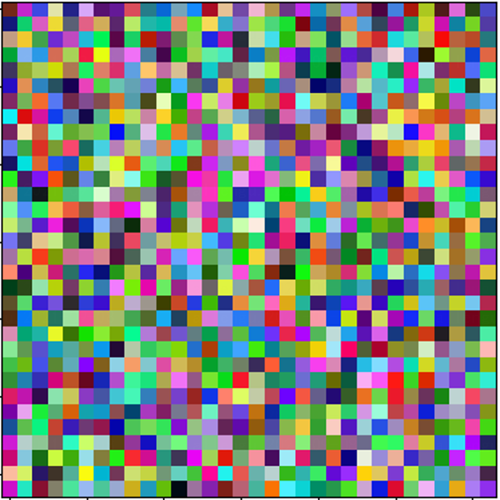}
    \subcaption*{$9.3\text{e-}2$}
  \end{subfigure}
  \hspace{-7mm}
  \vspace{1mm}
  
  \hspace{-9.8mm}
  \begin{subfigure}{0.2\textwidth}
    \centering
    \includegraphics[keepaspectratio=true, width=46pt]{{imgs/rec_img/ori.PNG}}
    \subcaption*{\textbf{$\text{BP}_\text{MSE}$}:0\\Original}
  \end{subfigure}
      \hspace{-10mm}
  \begin{subfigure}{0.2\textwidth}
    \centering
    \includegraphics[keepaspectratio=true, width=46pt]{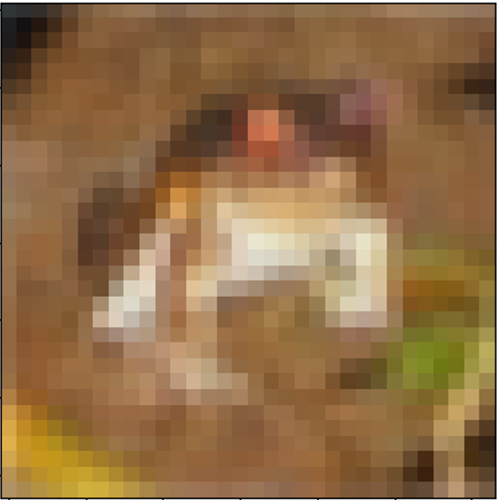}
    \subcaption*{$1.3\text{e-}3$\\DL}
  \end{subfigure}
    \hspace{-10mm}
  \begin{subfigure}{0.2\textwidth}
    \centering
    \includegraphics[keepaspectratio=true, width=46pt]{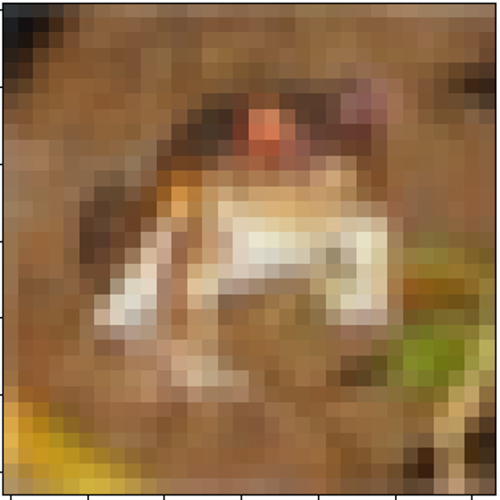}
    \subcaption*{$3.0\text{e-}3^{\Huge{*}}$\\DLDP-A}
  \end{subfigure}
     \hspace{-10mm}
  \begin{subfigure}{0.2\textwidth}
    \centering
    \includegraphics[keepaspectratio=true, width=46pt]{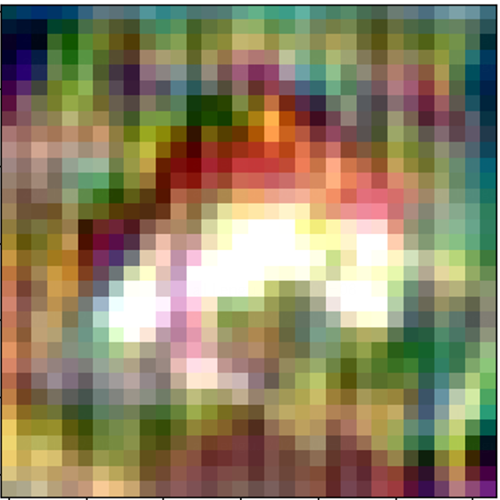}
    \subcaption*{$3.4\text{e-}2$\\DLDP-B}
  \end{subfigure}
      \hspace{-10mm}
  \begin{subfigure}{0.2\textwidth}
    \centering
    \includegraphics[keepaspectratio=true, width=46pt]{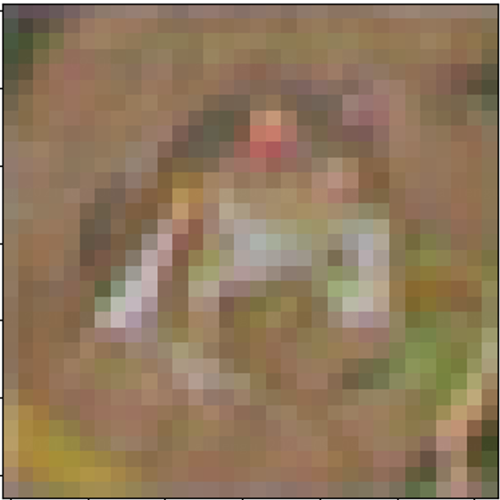}
    \subcaption*{$1.8\text{e-}2$\\SplitFed}
  \end{subfigure}
      \hspace{-10mm}
  \begin{subfigure}{0.2\textwidth}
    \centering
    \includegraphics[keepaspectratio=true, width=46pt]{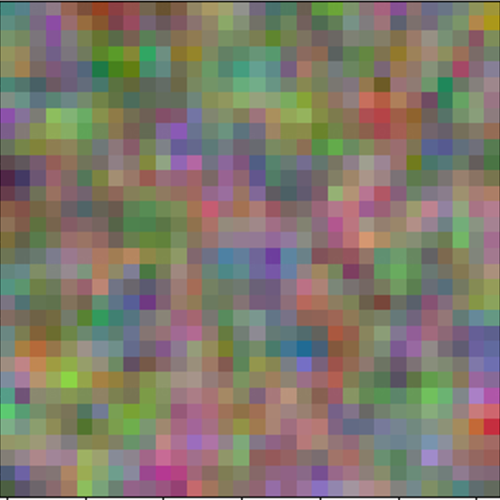}
    \subcaption*{$9.2\text{e-}2$\\FDL-PP}
  \end{subfigure}
    \hspace{-7mm}
    
 \hspace{-9.8mm}  
 \begin{subfigure}{0.2\textwidth}
  \centering
    \includegraphics[width=46pt]{{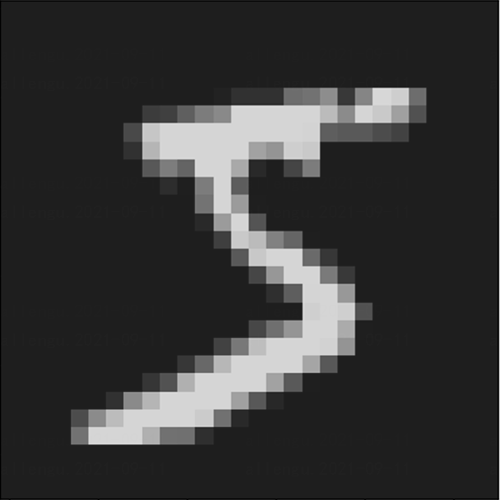}}
    \subcaption*{\textbf{$\text{BP}_\text{{MSE}}$}:0}
  \end{subfigure}
  \hspace{-10mm}
  \begin{subfigure}{0.2\textwidth}
  \centering
    \includegraphics[keepaspectratio=true, width=46pt]{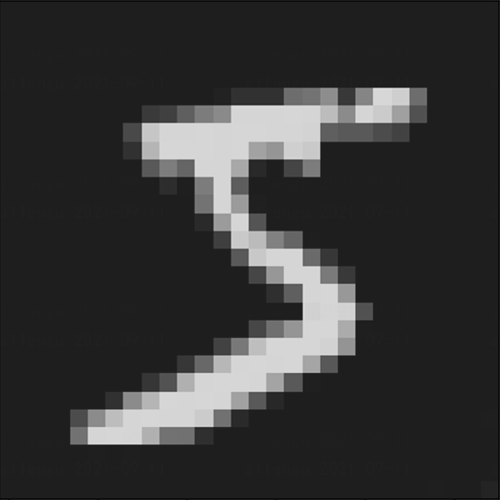}
    \subcaption*{$5.8\text{e-}6$}
  \end{subfigure}
    \hspace{-10mm}
  \begin{subfigure}{0.2\textwidth}
    \centering
    \includegraphics[keepaspectratio=true, width=46pt]{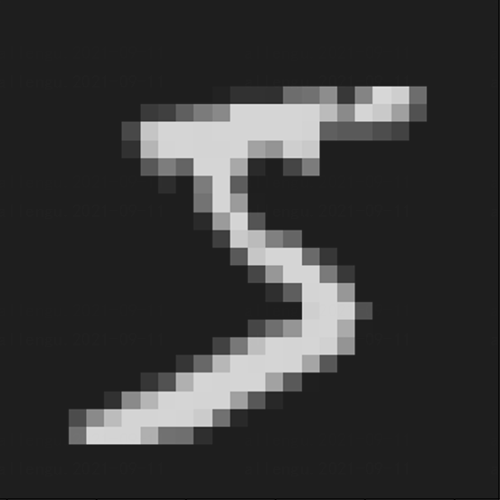}
    \subcaption*{$1.3\text{e-}5^{\Huge{*}}$}
  \end{subfigure}
    \hspace{-10mm}
  \begin{subfigure}{0.2\textwidth}
    \centering
    \includegraphics[keepaspectratio=true, width = 46pt]{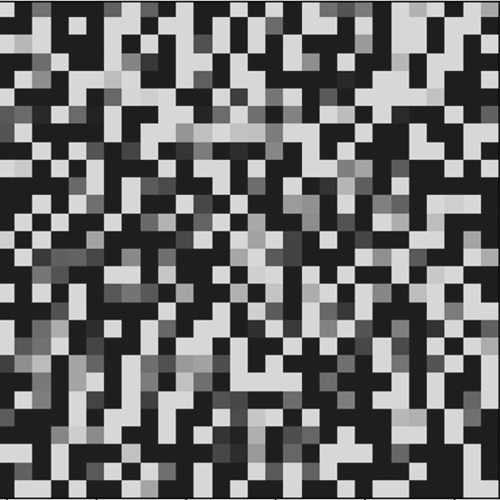}
    \subcaption*{$9.1\text{e-}2$}
  \end{subfigure}
    \hspace{-10mm}
  \begin{subfigure}{0.2\textwidth}
    \centering
    \includegraphics[keepaspectratio=true, width=46pt]{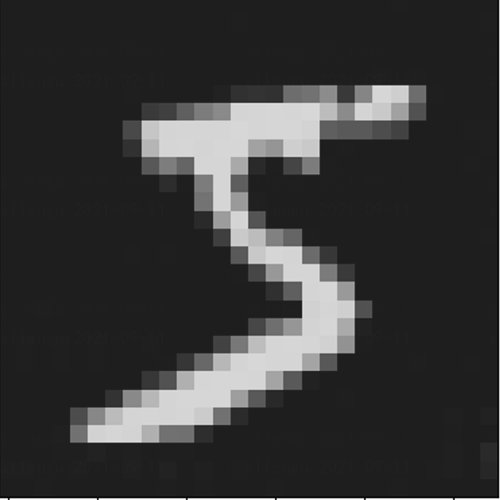}
    \subcaption*{$2.9\text{e-}5$}
  \end{subfigure}
    \hspace{-10mm}
  \begin{subfigure}{0.2\textwidth}
    \centering
    \includegraphics[keepaspectratio=true, width=46pt]{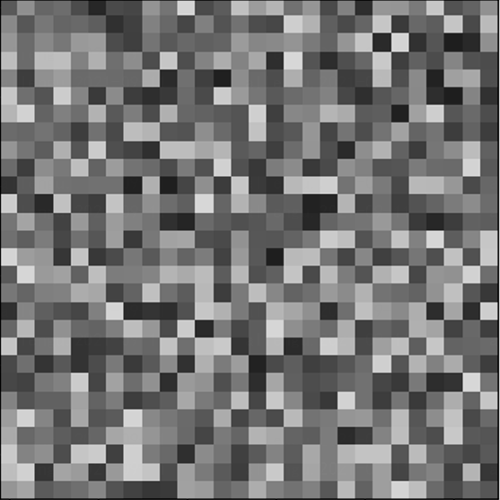}
    \subcaption*{$9.2\text{e-}2$}
  \end{subfigure}
  \hspace{-7mm}
  \vspace{1mm}
  
  \hspace{-9.8mm}
  \begin{subfigure}{0.2\textwidth}
    \centering
    \includegraphics[keepaspectratio=true, width=46pt]{{imgs/rec_img_mnist/ori.PNG}}
    \subcaption*{\textbf{$\text{BP}_\text{MSE}$}:0\\Original}
  \end{subfigure}
      \hspace{-10mm}
  \begin{subfigure}{0.2\textwidth}
    \centering
    \includegraphics[keepaspectratio=true, width=46pt]{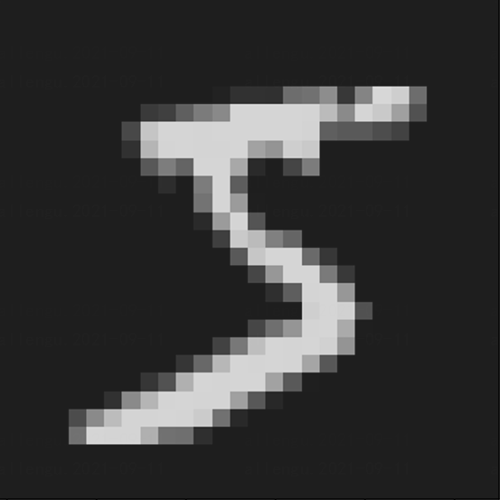}
    \subcaption*{$4.5\text{e-}3$\\DL}
  \end{subfigure}
    \hspace{-10mm}
  \begin{subfigure}{0.2\textwidth}
    \centering
    \includegraphics[keepaspectratio=true, width=46pt]{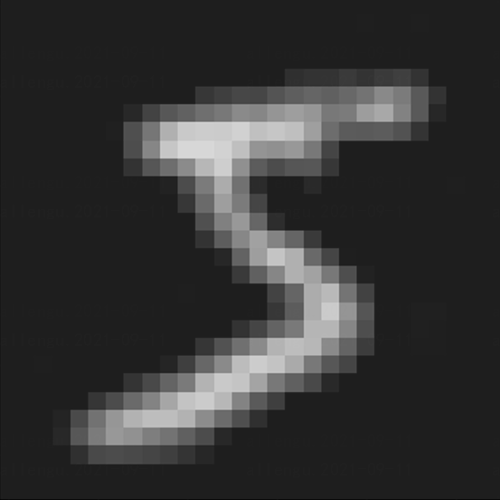}
    \subcaption*{$2.6\text{e-}2^{\Huge{*}}$\\DLDP-A}
  \end{subfigure}
     \hspace{-10mm}
  \begin{subfigure}{0.2\textwidth}
    \centering
    \includegraphics[keepaspectratio=true, width=46pt]{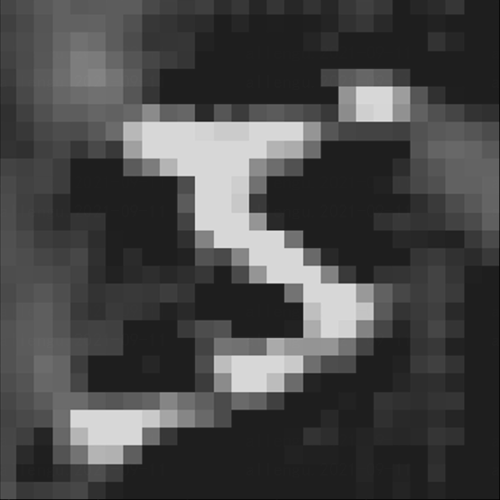}
    \subcaption*{$4.7\text{e-}2$\\DLDP-B}
  \end{subfigure}
      \hspace{-10mm}
  \begin{subfigure}{0.2\textwidth}
    \centering
    \includegraphics[keepaspectratio=true, width=46pt]{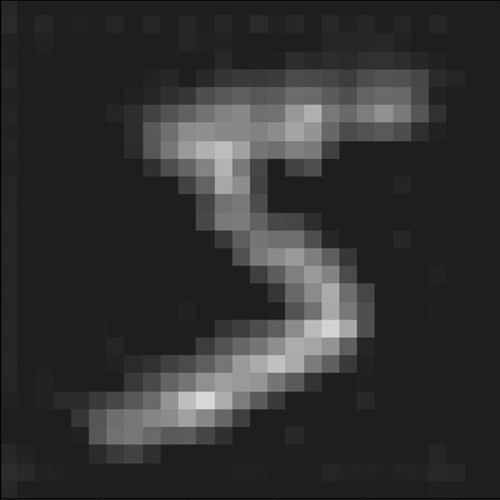}
    \subcaption*{$3.0\text{e-}2$\\SplitFed}
  \end{subfigure}
      \hspace{-10mm}
  \begin{subfigure}{0.2\textwidth}
    \centering
    \includegraphics[keepaspectratio=true, width=46pt]{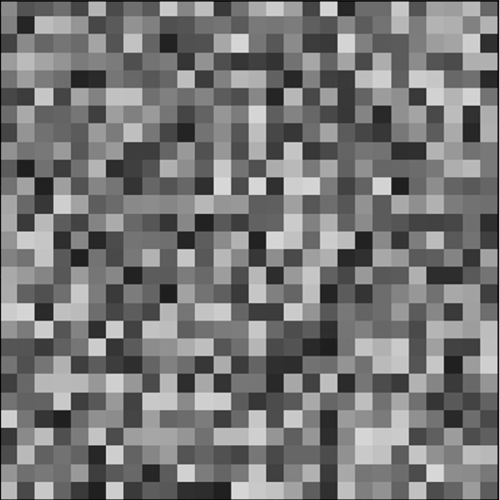}
    \subcaption*{$9.7\text{e-}2$\\FDL-PP}
    \end{subfigure}
    \hspace{-7mm}
    
\caption{\small Reconstructed images in different protection methods under Deep Leakage (top row) and Model Inversion attack (bottom row) in MNIST and CIFAR10. From left to right: original image, Distributed Learning (DL), protected with DLDP-A, DLDP-B, SplitFed and FDL-PP. 
  Bayesian Privacy guarantee (\textbf{$\text{BP}_\text{{MSE}}$}) are reported under each restored image (with * indicating BP guarantee provided by DLDP-A protection).} 
  \label{fig:rec_imgs}
    \vspace{-10pt}
\end{figure}

Table \ref{tab:compare-three-aspect} compares \textit{model performances}, \textit{privacy-preserving} capabilities in terms of Bayesian Privacy measured by both $\mathbf{BP_{KI}}$ and $\mathbf{BP_{MSE}}$, and protection method \textit{complexity} in terms of elapse time (s) for different protection methods. 
Model performances (accuracies) reported in \textbf{BOLD} are those with performances drop less than 1.5\% w.r.t. the best performance. 
Bayesian Privacy Guarantees reported in \textbf{BOLD} are those with restoration error ($\mathbf{BP_{MSE}}$) more than 0.05, ensuring sufficient privacy-preserving capabilities (see Fig. \ref{fig:rec_imgs} for example restored images). Bayesian Privacy losses reported in \textbf{BOLD} are those with restored data distribution similar to random distribution (with KL divergence no more than 0.3).   For training time reported in \textbf{BOLD} are those less than 100 seconds per epoch. 

It was observed that Bayesian Privacy guarantee (by MSE) faithfully reflected the quality of training images reconstructed by two attacking methods, regardless of different protection methods i.e. DLDP, SplitFed and FDL-PP being employed. Fig. \ref{fig:rec_imgs} shows example ground truth training images, corresponding reconstructed images and Bayesian Privacy measures. \textit{Bayesian Privacy leakages} also provides accurate estimations of information leaked from the reconstructed images, without using ground truth images.

It must be noted that, unlike the proposed Bayesian Privacy, Differential Privacy loss estimated by either the strong composition Theorem or moments accountants turn out to be too loose and unpractical in the face of two restoration attacks employed in experiments. For instance, in Fig. \ref{fig:rec_imgs}, the DLDP-A protection was unable to defeat Deep Leakage and Model inversion attacks whereas restored images were almost identical to the original image. In this case, Bayesian Privacy guarantee ($\text{BP}_\text{MSE}$) measured as low as $1.4\text{e-}6$ and $3.0\text{e-}3$ respectively. However, Differential Privacy still reported an acceptable level of privacy loss at  ($15.8,10^{-5}$)-\textit{differential privacy} with Gaussian noise ($\sigma=0.002$) being added \cite{abadi2016deep}.  

\begin{figure*}[t] 
\vspace{-10pt}         

\centering

\includegraphics[keepaspectratio=true, width=130pt]{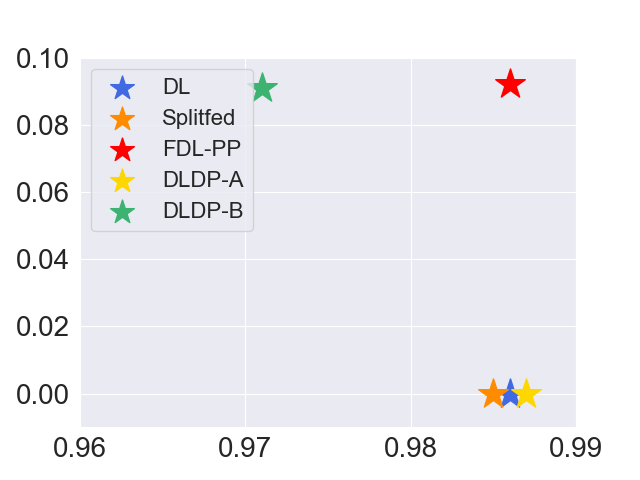}
\includegraphics[keepaspectratio=true, width=130pt]{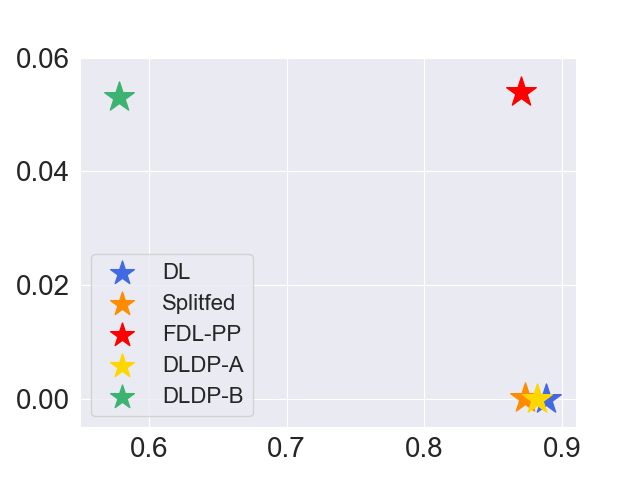}
\includegraphics[keepaspectratio=true, width=130pt]{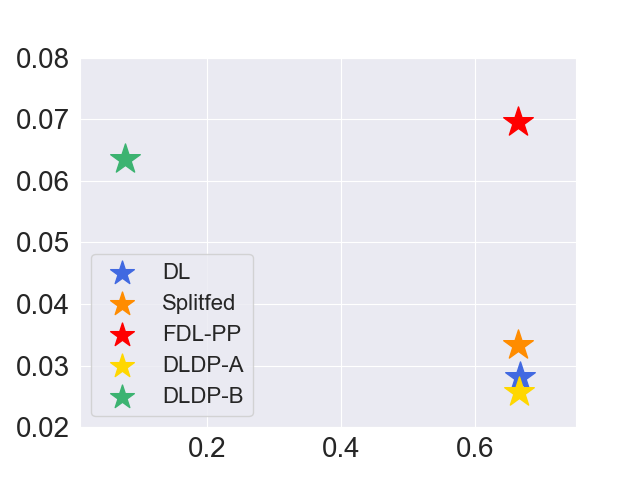}

\centering
\centering

\includegraphics[keepaspectratio=true, width=130pt]{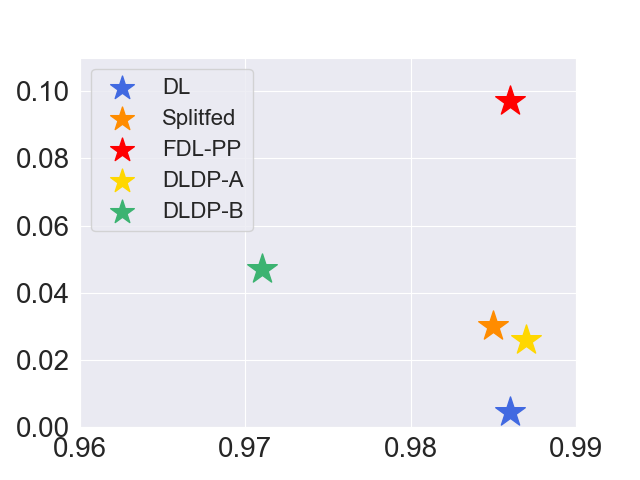}
\includegraphics[keepaspectratio=true, width=130pt]{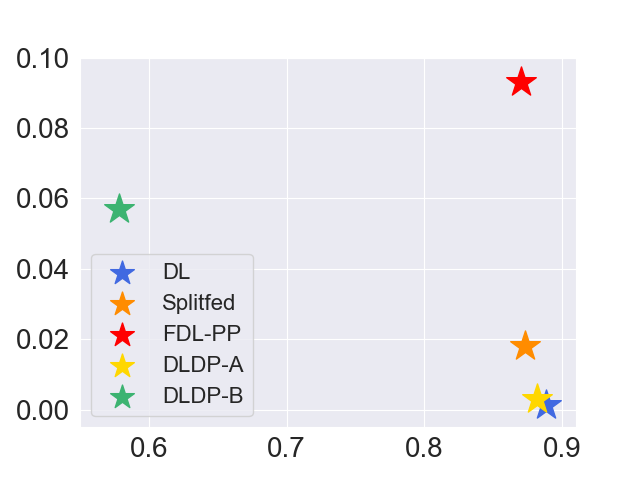}
\includegraphics[keepaspectratio=true, width=130pt]{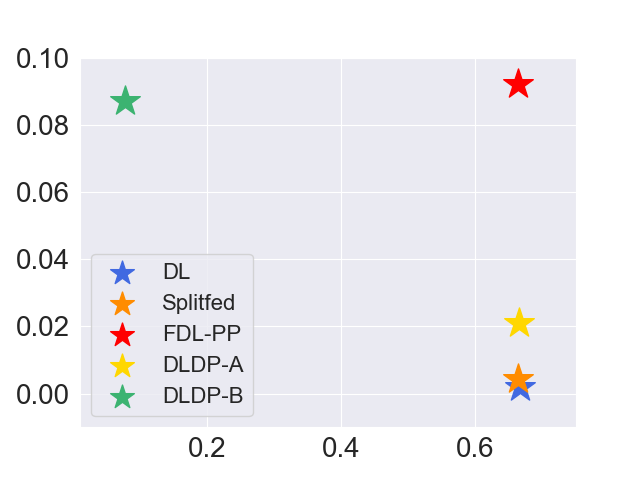}

\centering
\caption{Comparison of different methods in MNIST, CIFAR10 and CIFAR100 in terms of model performance and Bayesian Privacy Guarantee under deep leakage attack and model inversion attack. The top row is deep leakage attack and bottom row is model inversion attack. The x axis represents the model performance while y axis reflects privacy guarantee ($\mathbf{BP_{MSE}}$). Our proposed method FDL-PP (red star lies in top right position) could preserve the privacy and model performance simultaneously.}
\vspace{-5pt}
\label{fig:star}
\end{figure*}

The proposed \textbf{FDL-PP} method outperformed DL, DLDP-A and SplitFed in terms of Bayesian Privacy guarantee (MSE) by a margin of 0.05 or more.
On the other hand, FDL-PP achieved high model accuracy comparable to that of DL with classification accuracy drop around 0.2\%, 1.2\% and 0.3\% tested for MNIST, CIFAR10 and CIFAR100 datasets respectively. 
In terms of computational complexity, FDL-PP took only 55 seconds elapse time per epoch. In contrast, the running time of Homomorphic Encryption based \textbf{CryptoNet} \cite{gilad2016cryptonets} was orders of magnitudes higher than alternative protection approaches. Even with efficient improvements adopted for the \textbf{BatchCrypt} method \cite{batchCryp}, the running time was still twice as high as that of the proposed FDL-PP and at the cost of degraded model accuracy about 14\% lower for CIFAR10 classification.  
Among all methods compared in Table \ref{tab:compare-three-aspect} and Fig \ref{fig:star}, the proposed FDL-PP is the only protection method that \textit{simultaneously demonstrated satisfactory model performance, privacy preserving capabilities and low computational complexity}. Advantages of the proposed FDL-PP is also illustrated in Fig. \ref{fig:star}, in which FDL-PP is positioned on the top-right corner signalling simultaneous high model performances and privacy-preserving capabilities.

\begin{table*}[h!] 

\vspace{10pt}

\centering 
\caption{\label{tab:compare-three-aspect}   \footnotesize Comparison of different methods in terms of \textit{model performance} (accuracy), \textit{Bayesian Privacy Guarantee and loss}, and \textit{training time}, whereas results in \textbf{BOLD} are satisfactory (see text for explanation). "\textbackslash{}" for CryptoNet and BatchCrypt indicates no results measured. }
\tiny
\setlength{\tabcolsep}{0.2mm}
\renewcommand\arraystretch{2.0}
\begin{tabular}{|c|c|c|c|c|c|c|c|c|c|}
\hline

\multicolumn{3}{|c|}{}&  DL & DLDP-A   & DLDP-B  & SplitFed & CryptoNet & BatchCrypt & FDL-PP \\ \hline

\multirow{3}{*}{\textbf{\tabincell{c}{Model \\Performance}}}&\multicolumn{2}{c|}{MNIST}&\textbf{$\textbf{98.6\%}\pm{0.2\%}$} &$\textbf{98.7\%}\pm{0.1\%}$  &$97.1\%\pm{0.4\%}$ &$\textbf{98.5\%}\pm{0.2\%}$& $96.7\%\pm{0.3\%}$&  \textbackslash{}    &$\textbf{98.6\%}\pm{0.1\%}$ \\ \cline{2-10}
&\multicolumn{2}{c|}{CIFAR10}&$\textbf{88.8\%}\pm{4.6\%}$ &$\textbf{88.2\%}\pm{0.5\%}$  &$57.8\%\pm{1.5\%}$ &$\textbf{87.9\%}\pm{0.5\%}$ & \textbackslash{}      & $74.0\%$    &$\textbf{87.6\%}\pm{0.5\%}$ \\ \cline{2-10} 
&\multicolumn{2}{c|}{CIFAR100}&$\textbf{66.6\%}\pm{0.3\%}$ &$\textbf{66.3\%}\pm{0.4\%}$ & $7.65\%\pm{0.3\%}$& $\textbf{66.4\%}\pm{0.2\%}$& \textbackslash{}     & \textbackslash{}  &$\textbf{66.3\%}\pm{0.3\%}$  \\ \cline{1-10}

\multirow{6}{*}{\textbf{\tabincell{c}{Bayesian Privacy\\ Guarantee}}} &\multirow{2}{*}{MNIST}&DLG&$5.4\text{e-}6\pm{4.8\text{e-}10}$ &$1.3\text{e-}5\pm{1.2\text{e-}8}$ &$\textbf{9.1\text{e-}2}\pm{1.1\text{e-}2}$&$2.9\text{e-}5\pm{3.2\text{e-}10}$& \textbackslash{}&      \textbackslash{}     &$\textbf{9.2\text{e-}2}\pm{2.0\text{e-}2}$ \\ \cline{3-10}
&&MI&$4.5\text{e-3}\pm{4.0\text{e-}6}$ &$2.6\text{e-}2\pm{8.7\text{e-}5}$  &$4.7\text{e-}2\pm{9.1\text{e-}4}$&$3.0\text{e-}2\pm{7.9\text{e-}5}$& \textbackslash{}&      \textbackslash{}     &$\textbf{9.7\text{e-}2}\pm{4.5\text{e-}3}$\\ \cline{2-10}

&\multirow{2}{*}{CIFAR10}&DLG&$2.2\text{e-}8\pm{7.5\text{e-}11}$ &$1.4\text{e-}6\pm{7.0\text{e-}9}$ &$\textbf{5.3\text{e-}2}\pm{2.7\text{e-}3}$&$8.1\text{e-}5\pm{5.7\text{e-}8}$& \textbackslash{}&      \textbackslash{}     &$\textbf{5.4\text{e-}2}\pm{3.5\text{e-}3}$ \\ \cline{3-10}
&&MI&$1.3\text{e-3}\pm{5.0\text{e-}6}$ &$3.0\text{e-}3\pm{4.1\text{e-}5}$  &$3.4\text{e-}2\pm{1.7\text{e-}3}$&$1.8\text{e-}2\pm{1.0\text{e-}5}$& \textbackslash{}&      \textbackslash{}     &$\textbf{9.3\text{e-}2}\pm{3.5\text{e-}3}$\\ \cline{2-10}

&\multirow{2}{*}{CIFAR100}&DLG&$2.8\text{e-}2\pm{3.0\text{e-}3}$ &$2.8\text{e-}2\pm{2.0\text{e-}4}$ &$\textbf{6.4\text{e-}2}\pm{7.8\text{e-}3}$&$3.3\text{e-}2\pm{6.2\text{e-}3}$& \textbackslash{}&      \textbackslash{}     &$\textbf{7.0\text{e-}2}\pm{7.7\text{e-}3}$ \\ \cline{3-10}
&&MI&$1.9\text{e-3}\pm{4.1\text{e-}6}$ &$2.1\text{e-}3\pm{1.7\text{e-}5}$  &$\textbf{8.7\text{e-}2}\pm{8.1\text{e-}3}$&$2.3\text{e-}2\pm{1.5\text{e-}4}$& \textbackslash{}&      \textbackslash{}     &$\textbf{9.2\text{e-}2}\pm{9.1\text{e-}3}$\\ \cline{1-10}

\multirow{6}{*}{\textbf{\tabincell{c}{Bayesian Privacy\\Loss}}} &\multirow{2}{*}{MNIST}&DLG&$8.9\text{e-}1\pm{1.6\text{e-}4}$ &$8.2\text{e-}1\pm{1.2\text{e-}3}$&$\textbf{2.0\text{e-}1}\pm{2.4\text{e-}2}$&$8.9\text{e-}1\pm{2.2\text{e-}3}$ & \textbackslash{}&      \textbackslash{}     &$\textbf{1.6\text{e-}1}\pm{1.4\text{e-}2}$ \\ \cline{3-10}
&&MI&$8.2\text{e-}1\pm{5.7\text{e-}4}$ &$8.2\text{e-}1\pm{1.0\text{e-}3}$  &$7.8\text{e-}1\pm{1.1\text{e-}3}$ & $6.2\text{e-}1\pm{3.0\text{e-}2}$& \textbackslash{}&      \textbackslash{}     &$\textbf{2.8\text{e-}1}\pm{1.4\text{e-}2}$\\ \cline{2-10}

&\multirow{2}{*}{CIFAR10}&DLG&$5.3\text{e-}1\pm{3.4\text{e-}4}$ &$5.0\text{e-}1\pm{1.6\text{e-}3}$&$\textbf{2.9\text{e-}1}\pm{2.8\text{e-}3}$&$5.0\text{e-}1\pm{1.6\text{e-}4}$ & \textbackslash{}&      \textbackslash{}     &$\textbf{2.8\text{e-}1}\pm{1.7\text{e-}2}$ \\ \cline{3-10}
&&MI&$4.6\text{e-}1\pm{1.2\text{e-}3}$ &$4.6\text{e-}1\pm{1.6\text{e-}3}$  &$3.1\text{e-}2\pm{1.5\text{e-}2}$ &$4.2\text{e-}2\pm{8.4\text{e-}3}$& \textbackslash{}&      \textbackslash{}     &$\textbf{1.2\text{e-}1}\pm{8.2\text{e-}3}$  \\ \cline{2-10} 
&\multirow{2}{*}{CIFAR100}&DLG&$3.4\text{e-}1\pm{8.5\text{e-}3}$ &$3.3\text{e-}1\pm{1.6\text{e-}2}$&$\textbf{1.2\text{e-}1}\pm{1.2\text{e-}3}$&$3.1\text{e-}1\pm{5.0\text{e-}3}$ & \textbackslash{}&      \textbackslash{}     &$\textbf{1.1\text{e-}1}\pm{1.2\text{e-}2}$ \\ \cline{3-10}
&&MI&$4.3\text{e-}1\pm{1.0\text{e-}3}$ &$4.2\text{e-}1\pm{3.7\text{e-}3}$&$\textbf{1.4\text{e-}1}\pm{8.6\text{e-}3}$&$4.3\text{e-}1\pm{8.5\text{e-}3}$ & \textbackslash{}&      \textbackslash{}     &$\textbf{1.2\text{e-}1}\pm{1.0\text{e-}2}$   \\ \cline{1-10}

\multirow{2}{*}{\textbf{Training time (s)}} & \multicolumn{2}{c|}{MNIST}                   & \textbf{25.6}& \textbf{26.7} & \textbf{26.5}   & \textbf{30.1} & \multicolumn{1}{c|}{3009} &   \textbackslash{}       & \textbf{33.5} \\ \cline{2-10} 
                         &  \multicolumn{2}{c|}{CIFAR10}                  & \textbf{48.9}& \textbf{49.4} & \textbf{49.3}   & \textbf{52.9} &        \textbackslash{}                                    & \multicolumn{1}{c|}{$\sim$100} & \textbf{55.1} \\ \hline

\end{tabular}
\end{table*}
\newpage

\subsection{Empirical study of FDL-PP}

This section illustrates empirical justifications of three assumptions used in Theorem \ref{thm:pst} about the privacy-preserving capability of FDL-PP algorithm.

\textbf{Assumption 1: Convergence of private weights ($W$)}\\ 
In the case 1 of the Theorem \ref{thm:pst} (without passport), the Privacy Guarantees ($BP_{MSE}$) is limited by $||W-\tilde{W}||_2$, i.e. the difference of private weights between different clients. If $||W-\tilde{W}||_2$ decreases, the Privacy Guarantee is bounded from above by a small value. 

\begin{figure} [H]
    \centering
    \includegraphics[width=230pt] {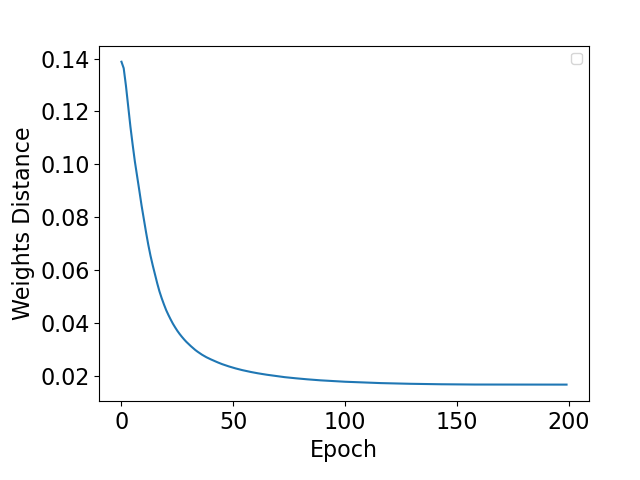}
    \caption{Differences between private layer weights of different client during the training process (Alexnet used for CIFAR10 datasets). }
    \label{fig:weight_dis}
\end{figure}

The Fig.\ref{fig:weight_dis} shows the change of average weights distance of private layer of different clients ($\sum\limits_{i,j=1}^N||W_i - W_j||_2^2/N^2$), where $W_i$ is private layer weight of $i_{th}$ client) in CIFAR10 during the training. It illustrates the difference of private weights among different clients decreases from 0.14 to 0.01 during training. Therefore, it is easy for attackers to restore the similar original input provided by the small Privacy Guarantee.

\begin{figure*}[t] 
\vspace{-10pt}         

\centering

 \begin{subfigure}{0.45\textwidth}
    \centering
    \includegraphics[keepaspectratio=true, width=200pt]{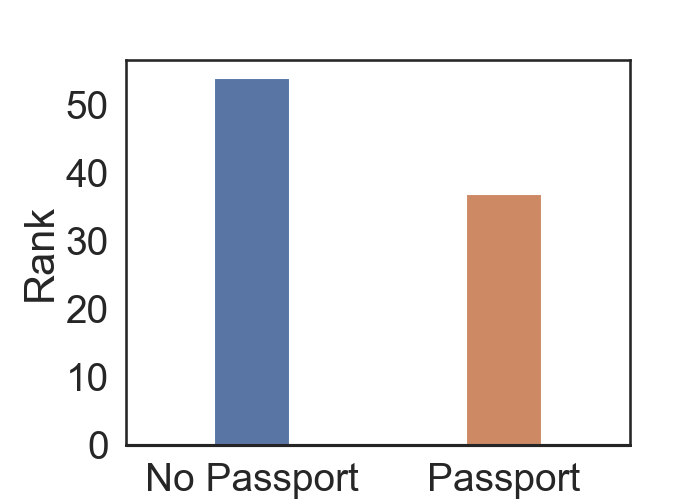}
    
  \end{subfigure}
 \begin{subfigure}{0.45\textwidth}
    \centering
    \includegraphics[keepaspectratio=true, width=200pt]{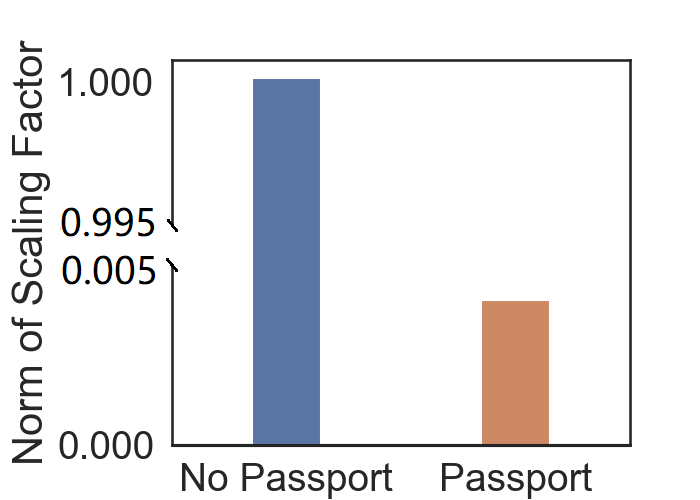}
  \end{subfigure}

\centering
\caption{Comparison of Passport and without Passport: After embedding with passport, the rank drops from 54 to 37, the scaling factor drops from 1.01 to 0.004}
\vspace{-5pt}
\label{fig:pst vs splitfed}
\end{figure*}

\textbf{Assumption 2: the rank of private weights decreases}\\
The case 2 of the Theorem \ref{thm:pst} (embedding passport) has one assumption that the rank of private layer weights decreases if the passport is used. The left figure of Fig. \ref{fig:pst vs splitfed} shows the rank of private weights decreases from 55 to 37, which verifies our assumption is reasonable.\\
\textit{Remark}: For the 2D matrix $W$, we take singular value decomposition of the $W$ and choose the $99\% \sum$(singular value) as threshold $\tau$. Finally, we use the number of singular values, which is larger than threshold, as the rank of $W$.

\textbf{Assumption 3: the norm of scaling factor decreases}\\
In the case 2 of the Theorem \ref{thm:pst} (embedding passport), the Privacy Guarantees ($BP_{MSE}$) has a lower bound $\frac{C}{||D_\gamma||_2 + ||\tilde{D_\gamma}||_2}$.

The right figure of Fig.\ref{fig:pst vs splitfed} demonstrates the scaling factor decreases from 1 to less than 0.005 after embedding passport. Therefore, the lower bound of eq. \ref{eq:pst} increases a lot, which provides a strong Bayesian Privacy Guarantee.

\section{Conclusion}

Recently proposed privacy attacks posed serious challenges to federated deep learning, in this work we proposed to reformulate the privacy definition from the lens of Bayesian data restoration and derived a rigorous Bayesian Privacy measure to quantify privacy leakage incurred by restored data. It was shown that estimated Bayesian Privacy leakages provided consistent accounts of the quality of restored images, and allowed us to investigate on the same ground a variety of protection mechanisms including randomization used in Differential Privacy.  Bayesian Privacy based theoretical analysis of different protection mechanisms constitute the first contribution of our work.  

Our second contribution lies in a novel passport-layer based protection method.  Due to the private passports fed to private passport layers in each client, private data can no longer be restored from exchanged information (see Theorem \ref{thm:pst}). Consequently, public neural network model parameters can be exchanged without encryption saving a great deal of computational complexity. This simple approach manifested itself, in extensive experiments with MNIST, CIFAR10 and CIFAR100 classifications, as a practical solution that simultaneously meets all three requirements in secure privacy protection, high model performance and low complexity. 

As for future work, we expect to investigate more protection and attacking methods in the generic Bayesian Privacy framework. In particular, to study adversarial protection methods that are aware of restoration attacks is one fruitful direction to explore further.

\newpage
\appendix
\section{Appendix A: Implementation details}
\subsection{Experiment Setting} This section illustrates the experiment settings of the empirical study on our Federated learning framework with Bayesian Privacy. 

\subsubsection{DNN Model Architectures} 

 The deep neural network architectures we investigated include the well-known AlexNet and ResNet-18. 
Private Passports $\mathbf{P}=(\mathbf{P}_\gamma, \mathbf{P}_\beta)$ are embedded into private convolution layers to derive crucial  normalization scale weights $\mathbf{W}^{\gamma}$ and shifting factor $\mathbf{W}^{\beta}$ 
 in LeNet, AlexNet and ResNet-18. Table \ref{tab: Net_arch} shows the detailed model architectures and parameter shape of LeNet, AlexNet and ResNet-18, which we employed in all the experiments.

\subsubsection{Dataset}
We evaluate classification tasks on standard MNIST, CIFAR10 and CIFAR100 dataset. The MNIST database of 10-class handwritten digits, has a training set of 60,000 examples, and a test set of 10,000 examples, CIFAR10 dataset consists of 60000 32x32 colour images in 10 classes, with 6000 images per class. CIFAR100 has 100 classes containing 600 images each. 
Respectively, we conduct stand image classification tasks of MNIST,CIFAR10 and CIFAR100 with LeNet, AlexNet and ResNet-18. 
We split the dataset into independently identically distribution (IID) setting for clients in horizontal federated learning experiments.

\subsubsection{FL Setting}
We simulate a $K=10$ clients horizontal federated learning system in a stand-alone machine 
with 8 Tesla V100-SXM2 32 GB GPUs and 72 cores of Intel(R) Xeon(R) Gold 61xx CPUs. 

The clients update the weight updates, server adopts \texttt{Fedavg}\cite{mcmahan2017communication} algorithm in to aggregate the model updates. The detailed experiment Hyper parameters we employ to conduct our federated learning
are listed in the Table \ref{tab:train-params}.

\subsection{Attack Algorithm}

We implement image restoration attack algorithms including Deep Leakage Attack and Model Inversion Attack. This section illustrates the algorithm implementation details of two type of attacks on Federated learning framework. 

\begin{table}[htbp]
\renewcommand{\arraystretch}{1.2}
    
    \setlength{\tabcolsep}{2.5mm}
    
	\begin{minipage}[t]{1\textwidth}
	    
	    \centering
		\begin{tabular}[c]{c|c|c|c}
			\toprule
			Layer name & Output size & Weight shape & Padding \\ 
			\hline
			Conv1 & 28 $\times$ 28 & 6 $\times$ 1 $\times$ 5 $\times$ 5 & 2 \\
			MaxPool2d & 14 $\times$ 14 & 2 $\times$ 2 &  \\
			Conv2 & 10 $\times$ 10 & 16 $\times$ 6 $\times$ 5 $\times$ 5 & 2 \\
			Maxpool2d & 5 $\times$ 5 & 2 $\times$ 2 &  \\
			Conv3 & 1 $\times$ 1 & 120 $\times$ 16 $\times$ 5 $\times$ 5 & 1 \\

			Linear & 84 & 84 $\times$ 120 &  \\
			Output & 10 & 10 $\times$ 84 &  \\\bottomrule
			
		\end{tabular} 
		\caption*{Modified LeNet Architecture} \label{tab: Lenet}
	\end{minipage}
    \vfill

	\begin{minipage}[t]{1\textwidth}
	    \centering
		\begin{tabular}[l]{c|c|c|c}
			\toprule
			Layer name & Output size & Weight shape & Padding \\ 
			\hline
			Conv1 & 32 $\times$ 32 & 64 $\times$ 3 $\times$ 5 $\times$ 5 & 2 \\
			MaxPool2d & 16 $\times$ 16 & 2 $\times$ 2 &  \\
			Conv2 & 16 $\times$ 16 & 192 $\times$ 64 $\times$ 5 $\times$ 5 & 2 \\
			Maxpool2d & 8 $\times$ 8 & 2 $\times$ 2 &  \\
			Conv3 & 8 $\times$ 8 & 384 $\times$ 192 $\times$ 3 $\times$ 3 & 1 \\
			Conv4 & 8 $\times$ 8 & 256 $\times$ 384 $\times$ 3 $\times$ 3 & 1 \\
			Conv5 & 8 $\times$ 8 & 256 $\times$ 256 $\times$ 3 $\times$ 3 & 1 \\
		
			MaxPool2d & 4 $\times$ 4 & 2 $\times$ 2 &  \\
			Linear & 10 & 10 $\times$ 4096 &  \\ \bottomrule
			
		\end{tabular} 
		\caption*{Modified AlexNet Architecture}\label{tab: AlexNet}
	\end{minipage}
    \vfill
	
	\begin{minipage}[t]{1\textwidth}
	    \centering

		\begin{tabular}[c]{c|c|c|c}
		
			\toprule
			Layer name & Output size & Weight shape & Padding \\ 
			\hline
			Conv1 & 32 $\times$ 32 & 64 $\times$ 3 $\times$ 3 $\times$ 3 & 1 \\ 
			\hline
			Res2 & 32 $\times$ 32 & $\begin{bmatrix}
				64 \times 64 \times 3 \times 3 \\
				64 \times 64 \times 3 \times 3 \\
			\end{bmatrix} \times 2$ & 1 \\
			\hline
			Res3 & 16 $\times$ 16 & $\begin{bmatrix}
				128 \times 128 \times 3 \times 3 \\
				128 \times 128 \times 3 \times 3 \\
			\end{bmatrix} \times 2$ & 1 \\
			\hline
			Res4 & 8 $\times$ 8 & $\begin{bmatrix}
				256 \times 256 \times 3 \times 3 \\
				256 \times 256 \times 3 \times 3 \\
			\end{bmatrix} \times 2$ & 1 \\
			\hline
			Res5 & 4 $\times$ 4 & $\begin{bmatrix}
				512 \times 512 \times 3 \times 3 \\
				512 \times 512 \times 3 \times 3 \\
			\end{bmatrix} \times 2$ & 1 \\
			\hline
		    Pooling & 1 $\times$ 1 & 4 $\times$ 4 &  \\
			\hline
			Linear & 100 & 100 $\times$ 512 &  \\ \bottomrule
		\end{tabular}
		\caption*{Modified ResNet-18 Architecture}\label{tab: Resnet18}
	\end{minipage} 
	\caption{Network Architecture adopted in Federated Learning with Bayesian Privacy }\label{tab: Net_arch}
\end{table}	

\begin{table*}[ht]
    \vspace{-10pt}
    \centering
    \renewcommand{\arraystretch}{1.2}
    \normalsize
	\resizebox{1.0\textwidth}{!}{
	    
		\begin{tabular}{c|ccc}
			\hline
			
			Hyper-parameter & LeNet &AlexNet & ResNet-18 \\ \hline
			Client numbers & 10 & 10 &10\\
			Private Passport Layers & Conv1 & Conv1 &  Conv1 \\
			Activation function & ReLU&ReLU & ReLU \\
			Optimization method & SGD&SGD & SGD \\
			Momentum & 0.9&0.9 & 0.9 \\
			Learning rate &0.01& 0.01 & 0.01 \\
			Batch size & 10&10 & 64 \\
			Data Distribution & IID & IID & IID \\
			Global Epochs & 100 & 200 &400\\
			Local Epochs & 1&1 & 1\\
			Learning rate decay & 0.99 at each global Epoch& 0.99 at each global Epoch &  0.99 at each global Epoch \\
		
			\hline
	\end{tabular}}
	\caption{Training parameters for Federated AlexNet$_{\mathbf{p}}$ and ResNet$_{\mathbf{p}}$-18, respectively ($\dagger$ the learning rate is scheduled as 0.01, 0.001 and 0.0001 between epochs [1-100], [101-150] and [151-200] respectively). }
	\label{tab:train-params}
	\vspace{-10pt}
\end{table*}

\subsubsection{Deep Leakage Attack:} 

Deep leakage attack(DLG)\cite{zhu2020deep} is implemented in the white box manner, server reconstructs client's private data only according to the federated model $W$ and gradients $\nabla W$ sent by clients. 
We refer to the source code in \url{https://github.com/mit-han-lab/dlg} and implement algorithm as shown in Alg. \ref{algorithm:deep_leakage}.  
The parameters of DLG attack please refer to Table \ref{tab:attack-params}

\begin{algorithm}[H]
    \caption{Deep Leakage Attack \cite{zhu2020deep}}\label{algorithm:deep_leakage}
    
     \textbf{Input:} 
        $F(\mathbf{x};W)$: Differentiable machine learning model;
        $W$: parameter weights; $\nabla W$: gradients calculated by training data;$\lambda$: tradeoff between prior and posteriori information
    \textbf{Output:} private training data $\mathbf{x}, \mathbf{y}$ 
    \begin{algorithmic}[1]
  
    \State Initialize $W_0$
	\For{$t$ in communication round $T$} \vspace{2pt}
	\State Send global model parameters $W_t$ to each clients
	\For{$i$ in clients $N$} \textbf {in parallel}
	\State $W_i^{t} \leftarrow$ \textbf{Local Training}
    \State Send public layer update $Pub(W_i^{t})$
    \State $Priv(W_i^{t}) \leftarrow$ \textbf{Local Private Training}
	\EndFor 
	\State \textbf{Information Leakage:}
    \State Leak $W$ and $\nabla W$ of victim client to server $S$
    \State \textbf{return} DLG($F$, $W$, $\nabla W$)
	\EndFor     
    \State \textbf{Attack:}
    \Procedure{DLG}{$F$, $W$, $\nabla W$} 
        \State $\mathbf{x'}_{1} \gets \mathcal{N}(0, 1)$ 
               , $\mathbf{y'}_{1} \gets \mathcal{N}(0, 1)$\\
        \Comment{Initialize dummy inputs and labels.}
        \For{$i \gets 1 \textrm{ to } n$}
                \State $
                    \nabla W'_i \gets 
                        \partial \ell (F(\mathbf{x'}_i, W_{t}), \mathbf{y'}_i)
                        /
                        \partial W_t 
                    $\\ \Comment{Compute dummy gradients.} 
                \State 
                    $ \mathbb{D}_i \gets ||\nabla W'_i - \nabla W||^2 + \lambda TV(x)$ 
                \State 
                    $\mathbf{x}'_{i+1} \gets \mathbf{x}'_i - \eta \nabla_{\mathbf{x}'_i} \mathbb{D}_i$
                   ,
                    $\mathbf{y}'_{i+1} \gets \mathbf{y}'_i - \eta \nabla_{\mathbf{y}'_i} \mathbb{D}_i$ \\
                    \Comment{Update data to match gradients.}
        \EndFor
        \State \textbf{return} $\mathbf{x}'_{n+1}, \mathbf{y}'_{n+1}$
    \EndProcedure
    \end{algorithmic}
\end{algorithm}

\begin{table}[H]
    \vspace{-10pt}
    \normalsize 
    \renewcommand{\arraystretch}{1.3}
    \setlength{\tabcolsep}{1.4mm}
    \centering
	\resizebox{0.48\textwidth}{!}{
		\begin{tabular}{c|cc}
			\hline
			Hyper-parameter & Value \\ \hline
			Optimization method & L-BFGS \cite{liu1989limited}\\
			Norm Type & L2  \\
			Learning rate of DLG & 0.001, 0.01 \\
			Attack Batchsize of DLG & 1 \\
			Learning rate of MIA & 0.001, 0.01 \\
			Attack Iterations of DLG & 300, 1000  \\
			Attack Iterations of MIA & 1500, 2000  \\
			\hline
	    \end{tabular}}\caption{Attack Parameters for Deep Leakage Attack and Model Inversion Attack}\label{tab:attack-params}
	    \vspace{-10pt}
\end{table}

\subsubsection{Model Inversion Attack:} 

Model Inversion Attack(MIA)\cite{he2019model} is implemented in the black-box setting of federated learning, that server $S$ conspires with client $C_1$ to attack the private data of client $C_0$, the algorithm is implemented as Alg. \ref{algorithm:mia_leakage}.

\begin{algorithm}[htbp]
    \caption{Model Inversion Attack \cite{he2019model}}\label{algorithm:mia_leakage}
    
     \textbf{Input:} 
        $F_{W_1}$: the model of attacker client $C_1$, 
        $F_{W_0}(x_0)$: the intermediate output of sensitive input $x_0$, $T$: maximum number of iterations, $\lambda$: tradeoff between prior and posteriori information;\\
    \textbf{Output:} 
    private training data $\mathbf{x}$ 
    \begin{algorithmic}[1]

    \State Initialize $W_0$
	\For{$t$ in communication round $T$} \vspace{2pt}
	\State Send global model parameters $W_t$ to each clients
	\For{$i$ in clients $N$} \textbf {in parallel}
	\State $W_i^{t} \leftarrow$ \textbf{Local Training}
    \State Send public layer update $Pub(W_i^{t})$
    \State $Priv(W_i^{t}) \leftarrow$ \textbf{Local Private Training}
	\EndFor 
	\State \textbf{Information Leakage:}
    \State Leak $F_{W_0}(x_0)$ of victim client $C_0$ to attacker $C_1$
    \State \textbf{return} MIA($F$, $W_1$, $F_{W_0}(x_0)$)
	\EndFor     
	\State \textbf{Attack:}
    \Procedure{MIA}{$F$, $W_1$, $F_{W_0}(x_0)$} 
        \State $L(x) = ||F_{W_1}(x) - F_{W_0}(x_0)||_2^2 + \lambda TV(x)$
        \For{$i \gets 1 \textrm{ to } T$}
                \State Minimize $L(x)$ with L-BFGS algorithm\\
                    \Comment{Update data to match outputs.}
        \EndFor
        \State \textbf{return} $x$
    \EndProcedure
    \end{algorithmic}
\end{algorithm}

\subsection{Privacy Preserving Algorithm}
We implement privacy-preserving FL algorithms including Differential Privacy, SplitFed and FDL-PP. This section illustrates the algorithm implementation details of our privacy preserving algorithm. 

\subsubsection{Federated Learning with Differential Privacy} 
  We implement Differential Privacy(DP)\cite{dwork2006calibrating} from the client's local perspective: during each local training round, each local client adds Gaussian noise\cite{geyer2017differentially} to the network weight updates before broadcasting to the central server. 
  
  The Gaussian noise we adopt obeys Gaussian distribution $\mathcal{N}(\mu, \sigma^2)$, we refer to the source code in  \url{https://github.com/cyrusgeyer/DiffPrivate_FedLearning}, and   implement algorithm as shown in Alg. \ref{alg:fl-dp}.

\begin{algorithm}[htbp]
	\caption{DL-DP Framework}
	\begin{algorithmic}[1]
	   \Statex \textbf{Input:} communication rounds $T$, client number $N$, number of local epochs $E$, Gaussian noise std $\sigma^2$

        \Statex \textbf{Output:} the final model $\Phi$ \vspace{4pt}
	    
	    \State Initialize $W_0$
		\For{$t$ in communication round $T$} \vspace{2pt}
		\State Send global model parameters $W_t$ to each clients
		\For{$i$ in clients $N$} \textbf {in parallel}
		\State $W_i^{t} \leftarrow$ \textbf{Local Training with DP}
	    \State Send weight update $W_i^{t}$

		\EndFor 
		\State \textbf{Server Update:}
        \State Aggregate $\{Pub(W_t^i)\}_{i=1}^{N}$ with $\texttt{FedAvg}$ algorithm
		\EndFor 
		
		\State \textbf{Local Training with DP:}
		\For{number of training iterations}
			\State Sample mini-batch $(X, Y)$
			\State Cross-entropy loss $L=L(X; Y)$
			\State Backpropagate using $L$ and update $W_i^{t}$
		
		\EndFor
		\State $W_i^{t} = W_i^{t} + \mathcal{N}(\mu, \sigma^2)$

	\end{algorithmic}\label{alg:fl-dp}
\end{algorithm}

\subsubsection{Splitfed with Passport}

Consider a neural network $\Phi(x;w,b):X \to R$, where $x\in X$, $w$ and $b$ are the weights and biases of neural networks, and $C$ is the output dimension. In the machine learning task, we aim to optimize $w$ and $b$ of network according to loss $L(\Phi(x;w,b))$,  where $x$ is the input
data and $y$ is the ground truth label. 

Inspired by \cite{thapa2020SplitFed}, we separate a deep neural network into several public and private layers as the 
shown in the following equation:
\begin{equation} \label{eq:public_private_all}
    \Phi(x;w,b) = \phi^{[1]} \circ \phi^{[2]} \circ \cdot \circ \phi^{[s]} \circ \psi^{[1]} \circ \psi^{[2]}  \cdots  \psi^{[t]}(x),
\end{equation} where $s$, $t$ are, respectively, the numbers of public and private layers.

We therefore further enhance the privacy preserving capability by adopting a private passport layer approach proposed by \cite{fan2021deepip, zhang2020passport} to protect DNN models. 
For each private convolution layer, a private passport layer consisting of a fully connected auto encoder (Encoder $E$ and Decoder $D$) and average pooling layer is used to derive the crucial normalization parameters $\gamma, \beta$. 
Note that private passports are used to compute these parameters as follows:

\begin{equation}
\begin{split}
        \mathbf{O}^l(\mathbf{X}_p) = &
        \gamma^l(\mathbf{W}_p^l* \mathbf{X}_c^l) + \beta^l, \\
         \gamma^l=&\text{Avg}\Big( \mathbf{D}\big(\mathbf{E}(\mathbf{W}_p^l* \mathbf{P}_\gamma^l)\big)\Big)\\
        \beta^l = &\text{Avg}\Big( \mathbf{D}\big(\mathbf{E}(\mathbf{W}_p^l* \mathbf{P}_\beta^l)\big)\Big)
\end{split}
\end{equation}
where * denotes the convolution operations, $l$ is the layer number, 
$\mathbf{X}_c$ is the private data fed to the convolution layer. 
$\mathbf{O}()$ is the corresponding linear transformation of
outputs, while $\mathbf{P}^l_\gamma$ and $\mathbf{P}^l_\beta$
are \textit{passports private to each client} used to derive scale factor $\gamma$ and bias term $\beta$ respectively. 

Scale factor $\gamma$ and bias term $\beta$ are enforced to take values depending on passports $\mathbf{P}^l_\gamma$ and $\mathbf{P}^l_\beta$ and this constraint can be succinctly represented by a regularization term to be minimized as follows:
\begin{equation}\label{eq:passport-constr}
\begin{split}
L_P(w, P) := &\| \gamma^l - \text{Avg}\Big( \mathbf{D}\big(\mathbf{E}(\mathbf{W}_p^l* \mathbf{P}_\gamma^l)\big)\Big) \|_2  + \\ 
&\| \beta^l - \text{Avg}\Big( \mathbf{D}\big(\mathbf{E}(\mathbf{W}_p^l* \mathbf{P}_\beta^l)\big)\Big)  \|_2,
\end{split}
\end{equation}
in which $w=\{ \mathbf{W}_p^l, \gamma^l, \beta^l \}$ is the neural network model parameters and $P = \{\mathbf{P}_\gamma^l,\mathbf{P}_\beta^l\}$ the passport for brevity. 

Combined with the \textit{cross-entropy} loss for the classification task, the NN model parameters $w=\{ \mathbf{W}_p^l, \gamma^l, \beta^l\}$ are therefore sought by minimizing the following loss:
\begin{equation}\label{eq:CE-passport}
\begin{split}
L_{\text{FP}}(&w,x,P)=L_{CE}(w, x)+L_P(w, P), \\
&w^* = \argmin_{w} L_{\text{FP}}(w,x,P), 
\end{split}
\end{equation}
in which $x$ is the given training data and $P$ the passport defined above. 

It must be noted that the regularization term $L_P(w, P)$ enforces the scale and bias term $\{\gamma, \beta\}$ and $w$ to take specific values that depend on not only the training data but also the passport. It is this modulation of NN parameters $w$ through the passport $P$ that plays an indispensable role in protecting the private model parameters and training data from being disclosed by restoration attacks (see detailed analysis in Appendix B, C and D).  

The training procedure of FDL-PP is illustrated in following three steps (shown in Algorithm \ref{alg:fdl-pp}):
\begin{itemize}
    \item Firstly, all clients carry out the forward and backward propagation on a global model in parallel and upload their 
    public layers of the model to the server;

    \item Secondly, the central server process the FederatedAverage \cite{mcmahan2017communication} of the public layers and update;
    
    \item Finally, clients download the updated public model, followed by back-propagation and updating to their  private layers with embedding the passport.
\end{itemize}

\begin{algorithm}[H]
    
	\caption{FDL-PP Framework}

	\begin{algorithmic}[1]
	   \Statex \textbf{Input:} communication rounds $T$, client number $N$, number of local epochs $E$, each client $i$ with its own passport tuple $(P_\gamma, P_\beta)$ 

        \Statex \textbf{Output:} the final model $\Phi$ \vspace{4pt}
	    
	    \State Initialize $W_0$
		\For{$t$ in communication round $T$} \vspace{2pt}
		\State Send global model parameters $W_t$ to each clients
		\For{$i$ in clients $N$} \textbf {in parallel}
		\State $W_i^{t} \leftarrow$ \textbf{Local Training}
	    \State Send public layer update $Pub(W_i^{t})$
	    \State $Priv(W_i^{t}) \leftarrow$ \textbf{Local Private Training}
		\EndFor 
		\State \textbf{Server Update:}
        \State Aggregate $\{Pub(W_t^i)\}_{i=1}^{N}$ with $\texttt{FedAvg}$ algorithm
		\EndFor 
		
		\State \textbf{Local Training:}
        
		\State  Initialize passport keys  $(P^l_\gamma, P^l_\beta)$ for targeted passport layers $N_{pass}$

		\For{number of training iterations}
			\State Sample mini-batch $(X, Y)$
			\For{$l$ in $N_{pass}$}
				\State Cross-entropy loss $L=L(X;P^l_\gamma, P^l_\beta; Y)$
			\EndFor
		
			\State Backpropagate using $L$ and update $W_i^{t}$
		\EndFor

		\State \textbf{Local Private Training:}
        \For{number of training iterations}
            \State Sample mini-batch $(X, Y)$
            \State Compute cross-entropy loss $L=L(X;P^l_\gamma,P^l_\beta; Y)$
			\For{layer $l$ in Public layers set $L_{pub}$}
				\State Frozen model parameters in layer $l$
			\EndFor
			\State Backpropagate using $L$ and update $Priv(W_i^{t})$	
		\EndFor
	\end{algorithmic}
\end{algorithm}

\subsection{Privacy Metrics}
For each images including the original images and reconstructed images, we firstly scale the original images and reconstructed images to the range (0,1).
For Bayesian Privacy guarantee, we calculate the mean square error between the original images and reconstructed images. For Bayesian Privacy loss, we statistics the numbers of pixel value of images in k intervals ($\frac{i}{k}, \frac{i+1}{k}$, $0\leq i \leq k-1$) as the distribution of reconstructed images, and view the uniform distribution as the distribution of no prior. The KL distance between these two distributions are the Bayesian Privacy loss.

\newpage
\bibliographystyle{IEEEtranS}
\bibliography{appendix.bib}

\end{document}